\documentclass[journal]{IEEEtai}

\usepackage{booktabs} 
\usepackage{authblk}
\usepackage{times}
\usepackage{amssymb}
\usepackage{multirow,booktabs,color,soul,threeparttable,rotating}
\usepackage{amsmath}
\usepackage{epsfig,amssymb,color}
\usepackage{algorithmic}
\usepackage{caption}
\usepackage{float}
\usepackage{mathrsfs}
\usepackage{color}
\usepackage{cite}
\usepackage{booktabs}
\usepackage{paralist}
\usepackage{epstopdf}

\usepackage{tcolorbox}
\definecolor{Gray}{gray}{0.85}
\definecolor{LightGray}{gray}{0.9}
\usepackage[ruled, vlined, linesnumbered]{algorithm2e}

\usepackage{graphicx, subfigure}

\graphicspath{{figures/}}

\begin{document}

\title{SoloGAN: Multi-domain Multimodal Unpaired Image-to-Image Translation via a Single Generative Adversarial Network}

\author{Shihua~Huang,
        Cheng~He,
        and Ran~Cheng,~\IEEEmembership{Senior Member, IEEE}
\thanks{S. Huang and R. Cheng are with the Research Institute of Trustworthy Autonomous Systems, Southern University of Science and Technology, Shenzhen 518055, China. They are also with the Guangdong Key Laboratory of Brain-Inspired Intelligent Computation, Department of Computer Science and Engineering, Southern University of Science and Technology, Shenzhen 518055, China. E-mail: shihuahuang95@gmail.com, ranchengcn@gmail.com (\textit{Corresponding author: Ran Cheng}).}
\thanks{C. He was with the Guangdong Provincial Key Laboratory of Brain-inspired Intelligent Computation, Department of Computer Science and Engineering, Southern University of Science and Technology, Shenzhen 518055, China. He is currently with the School of Electrical and Electronic Engineering, Huazhong University of Science and Technology, Wuhan 430074, China. Email: chenghehust@gmail.com.}
\thanks{This work was supported by the National Natural Science Foundation of China (No. 61906081 and U20A20306), the Shenzhen Science and Technology Program (No. RCBS20200714114817264), the Guangdong Provincial Key Laboratory (No. 2020B121201001), and the Program for Guangdong Introducing Innovative and Entrepreneurial Teams (Grant No. 2017ZT07X386).}
}
\markboth{IEEE Transactions on Artificial Intelligence, Vol. 00, No. 0, Month 2020}
{Huang \MakeLowercase{\textit{et al.}}: SoloGAN}

\maketitle

\begin{abstract}
Despite significant advances in image-to-image (I2I) translation with generative adversarial networks (GANs), it remains challenging to effectively translate an image to a set of diverse images in multiple target domains using a single pair of generator and discriminator. Existing I2I translation methods adopt multiple domain-specific content encoders for different domains, where each domain-specific content encoder is trained with images from the same domain only. Nevertheless, we argue that the content (domain-invariance) features should be learned from images among all of the domains. Consequently, each domain-specific content encoder of existing schemes fails to extract the domain-invariant features efficiently. To address this issue, we present a flexible and general SoloGAN model for efficient multimodal I2I translation among multiple domains with unpaired data. In contrast to existing methods, the SoloGAN algorithm uses a single projection discriminator with an additional auxiliary classifier and shares the encoder and generator for all domains. Consequently, the SoloGAN can be trained effectively with images from all domains such that the domain-invariance content representation can be efficiently extracted. Qualitative and quantitative results over a wide range of datasets against several counterparts and variants of the SoloGAN demonstrate the merits of the method, especially for challenging I2I translation datasets, i.e., datasets involving extreme shape variations or need to keep the complex backgrounds unchanged after translations. Furthermore, we demonstrate the contribution of each component in SoloGAN by ablation studies.
\end{abstract}

\begin{IEEEImpStatement}
Image-to-image (I2I) translation is a popular technology in image synthesis. Owing to the rapid development of artificial intelligence, particularly deep learning, I2I translation has been applied in a wide range of real-world scenarios such as image super-resolution, style transfer, and face swapping. Nevertheless, most existing methods require multiple pairs of generators and discriminators for diverse translations among multiple domains, which is neither efficient nor effective. Moreover, those methods train the content encoder for each domain independently, although the global information can only be learned with images from all domains. To address these issues in I2I translation among multiple domains, we propose a simple yet effective method. In a user study for ${cat} \rightarrow {dog}$ dataset, 93.65\% of users prefer our synthetic images, significantly outperforming the compared peer methods. The proposed method can benefit users in various industries including fashion design, image design, and movie production.
\end{IEEEImpStatement}

\section{Introduction}
\begin{figure*}[t]
\centering
\centering
\subfigure[Unimodal I2I translations]{
    \begin{minipage}[t]{0.335\linewidth}
        \centering
        \includegraphics[width=2.31in]{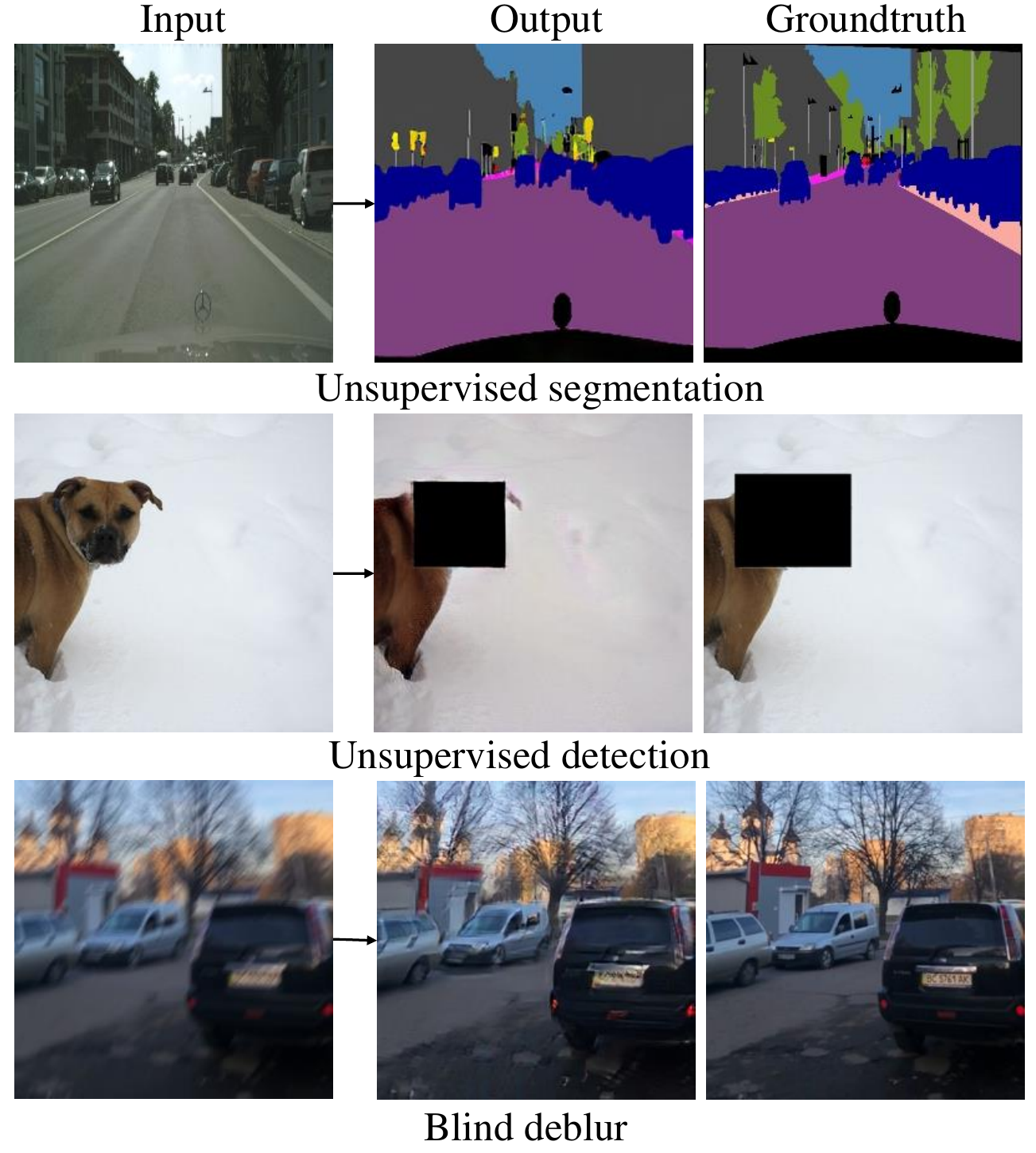}
    \end{minipage}%
}
\subfigure[Multimodal I2I translations]{
    \begin{minipage}[t]{0.63\linewidth}
        \centering
        \includegraphics[width=4.53in]{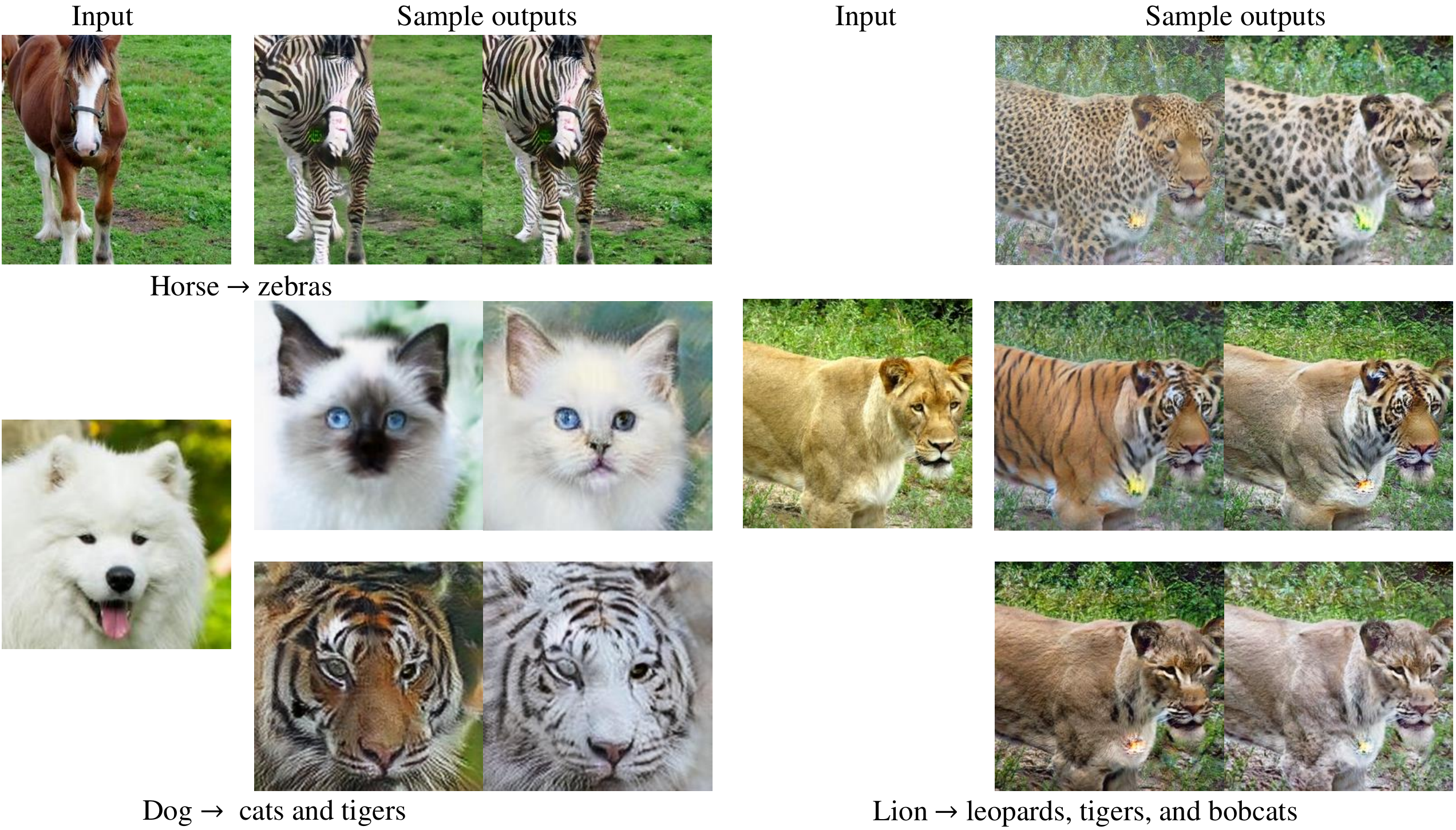}
    \end{minipage}%
}
\caption{Translated results achieved by (a) unimodal variant of SoloGAN and (b) SoloGAN.}
\label{fig:f1_results}
\end{figure*}

\begin{figure*}[!htbp]
\centering
\subfigure[Existing multimodal I2I translation methods]{
    \begin{minipage}[t]{0.32\linewidth}
        \centering
        \includegraphics[width=2.56in]{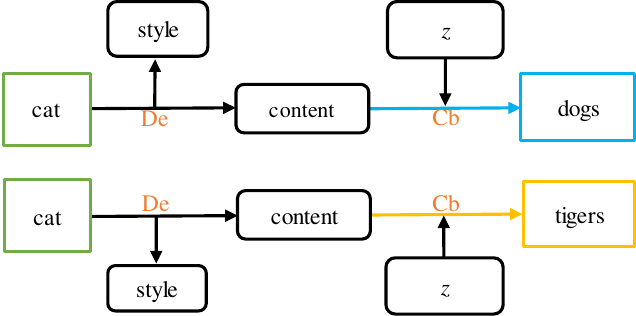}
    \end{minipage}%
}%
\subfigure[Our method]{
    \begin{minipage}[t]{0.5\linewidth}
        \centering
        \includegraphics[width=2.912in]{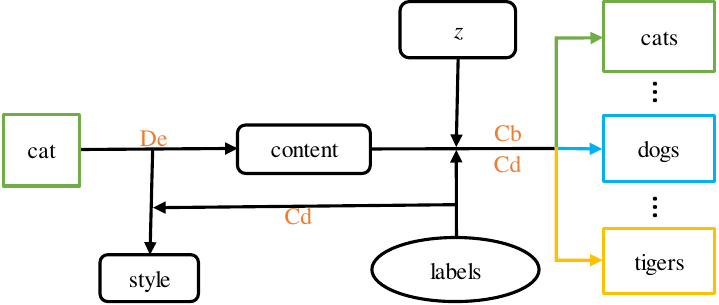}
    \end{minipage}%
}%
   \caption{Workflows of (a) existing multimodal I2I translation methods and (b) our method. De, Cb, and Cd represent Decompose, Combine, and Condition, respectively.
    To translate an image from a source domain into a target domain, the latent space of the image is first decomposed into the content and style.
A generator combines the content and different style vectors ($\mathbf{z}$, the random samples from $N(0,1)$) to generate different target images.
Taking the translation from the same cat into dogs and tigers as an example, existing methods need to setup two independent pairs of translations: cat to dog and cat to tiger.
In contrast, our method can obtain objects of different domains with the same encoder and generator when given different target domain labels.}
\label{fig:ill_intro}
\vspace{-0.2cm}
\end{figure*}

\IEEEPARstart{I}{mage-to-image} (I2I) translation aims to learn a function that changes domain-specific part/style of given image to target while preserving its domain-invariance part/content~\cite{MUNIT, DRIT}.
A variety of vision and graphics problems, e.g. semantic segmentation, object detection, and deblur, can be formulated as I2I translation problems [Fig.~\ref{fig:f1_results}(a)].
Generative adversarial networks (GANs)~\cite{GANs} have received extensive attention in recent years, and a number of GAN-based methods have been developed for vision tasks, including person reidentification~\cite{SPGAN, MPR}, superresolution~\cite{SRGAN, SR_GAN}, text-to-image synthesis~\cite{StackGAN, AttnGAN}, and facial attribute manipulation~\cite{AttGAN, GANimation, StarGAN}.
Significant advances have been made in I2I translation with the help of GANs~\cite{Pix2Pix, GatedGAN, AsymmetricGAN, PAN, SRA}. Among them, Pix2Pix~\cite{Pix2Pix} trains a conditional GAN with paired training data for supervised I2I translations. In contrast, some works attempt to learn \emph{unpaired I2I translation}\footnote{In this work, we refer to I2I translation without paired ground-truth supervision as \emph{unpaired I2I translation}}~\cite{DiscoGAN, DualGAN, CycleGAN, UNIT}.
For instance, the CycleGAN~\cite{CycleGAN} method introduces the cycle consistency loss for unpaired I2I translation. To further enhance the capability of CycleGAN in identifying the most discriminative foregrounds, some attention-guided generators are proposed in~\cite{AttentionGAN}, and a shared knowledge module along with absolute consistency loss is used in~\cite{GRGAN}.

Although much progress has been made, those methods cannot translate a single image to a set of diverse images in a target domain, which is known as multimodal I2I translation~\cite{BicycleGAN}.
Fig.~\ref{fig:f1_results}(b) shows one example of the multimodal I2I translation, where a horse is translated to a few zebras but with diverse appearances.
To address this issue, a number of multimodal I2I translation algorithms have been proposed in recent years, including MUNIT~\cite{MUNIT} and DRIT~\cite{DRIT}, where the main idea is to employ a pair of content and style encoders to embed an image from a specific domain into a domain-invariant space and a domain-specific space and then use a generator to map the latent codes to diverse outputs in a target domain [Fig.~\ref{fig:ill_intro}(a)].
However, those methods, including the improved GR-GAN~\cite{GRGAN}, are less efficient since multiple pairs of content and style encoders and GANs are required for multimodal I2I translation among multiple domains. An illustrative example to show the framework of existing methods for translating cat to dog and tiger is displayed in Fig.~\ref{fig:ill_intro}(a), where two independent cat content and style encoders, one dog and tiger generator, as well as one dog and tiger discriminators are involved.
These approaches use different domain-specific content encoders to learn domain-invariant features for different domains and train the model of each domain using the images from the same domain independently. 

\begin{figure}[tbp]
\centering
\vspace{-0.1cm}
\begin{tabular}{cc}
        \includegraphics[width=.45\linewidth]{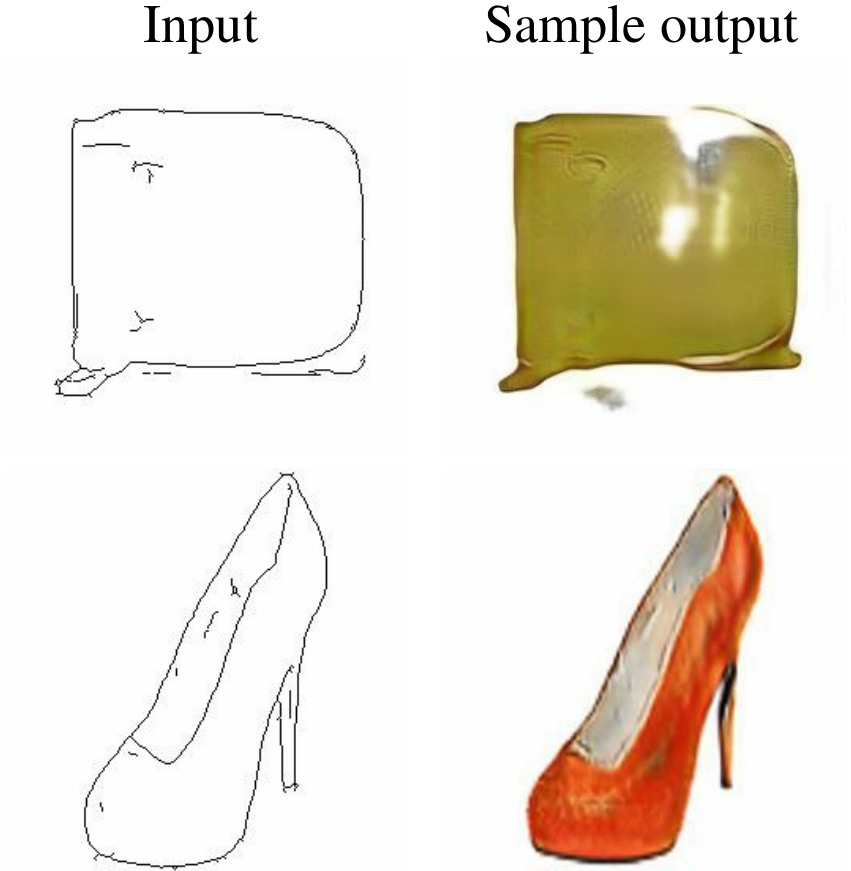} &
        \includegraphics[width=.45\linewidth]{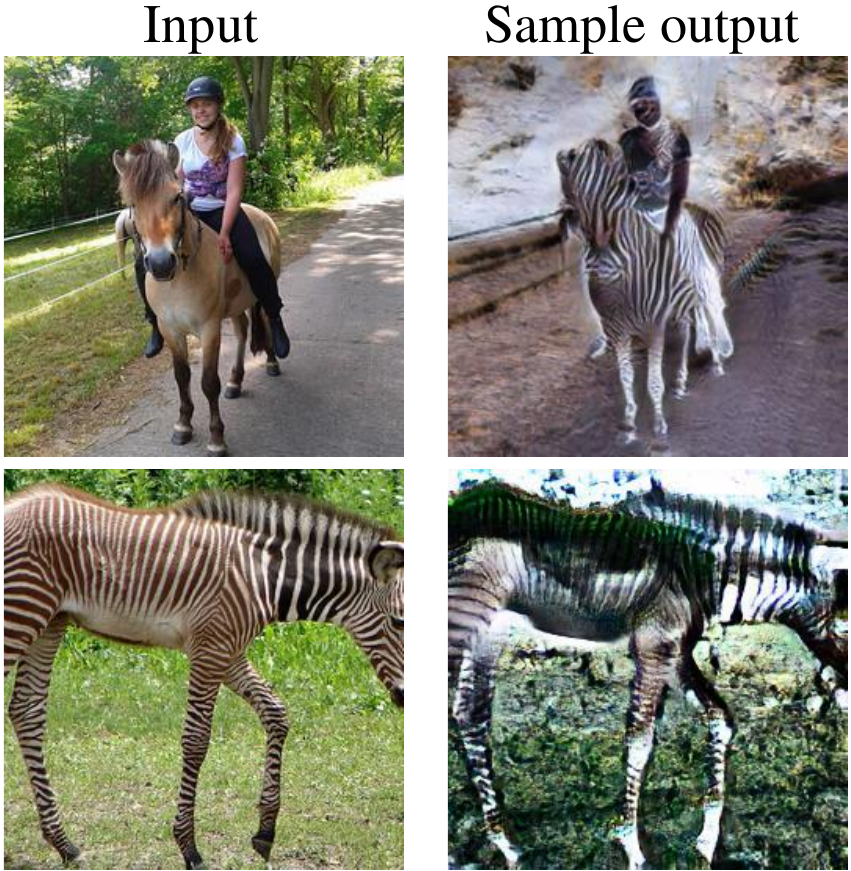} \\
           (a) $edges \rightarrow {bag}$ and $shoe$ & (b) $horse \leftrightarrow zebra$
           \end{tabular}
   \caption{(a) and (b) I2I translations with simple and complex backgrounds achieved by MUNIT~\cite{MUNIT}, respectively. MUNIT performs well in the simple backgrounds but fails in the complex backgrounds.}
\label{fig:munit_examples}
\vspace{-0.3cm}
\end{figure}

These approaches 
need to decompose the cat twice with two totally independent encoders, aiming to extract the domain-invariance among the domains of cat, dog, and tiger.
Nevertheless, the domain-invariance features should be learned from images from all domains.
It is unlikely for these schemes to translate objects of different scales with complex backgrounds effectively.
Consequently, existing methods can perform well in translating objects with simple backgrounds [as shown in Fig.~\ref{fig:munit_examples}(a)] but fail to translate objects with diverse appearances and complex backgrounds (e.g. translation from a horse in the wild to a zebra and vice versa, as shown in Fig.~\ref{fig:munit_examples}(b)), since the translation methods should be aware of the diverse backgrounds, which are also the domain-invariant components, among different domains.

To summarize, there are three main challenges to the existing  methods when performing multi-domain multimodal I2I translations:
\begin{itemize}
    \item \textbf{Diverse outputs:} Existing methods benefit from the merit of using a unified model to facilitate learning, but they often fail to generate diverse outputs of the same domain when given one input.
    \item \textbf{Multiple GANs:} Existing  methods benefit from the disentangled representation of content and style, but they often require multiple pairs of generator and discriminator for translation among multiple domains and need to train the content encoder independently with domain-specific images.
    \item \textbf{Ineffectiveness:} Existing methods are often ineffective in dealing with datasets that require \emph{extreme shape variations} or retain the \emph{complex background} after translation.  
\end{itemize}

To meet the above challenges, we propose SoloGAN for multi-domain multimodal I2I translation.
A single content encoder is used to encode the domain-invariant features of all the images from multiple domains; the style encoder and generator are shared among different domains in a conditional manner.
In addition, a projection discriminator~\cite{ProjectionD} with an additional auxiliary classifier is constructed instead of constructing multiple discriminators for different domains, e.g. IntersectGAN~\cite{IntersectGAN}.
\setlength{\tabcolsep}{1pt}

\begin{table*}[!htbp]
\centering
\caption{Comparisons of existing GAN-based I2I translation methods.}
\label{tb:comparisions}
\footnotesize
\begin{tabular}{ccccccccc}
\toprule
Method &Pix2Pix~\cite{Pix2Pix} &CycleGAN~\cite{CycleGAN} &StarGAN~\cite{StarGAN} &BicycleGAN~\cite{BicycleGAN} &MUNIT~\cite{MUNIT} &DRIT~\cite{DRIT} &SingleGAN~\cite{SingleGAN} &SoloGAN \\
\midrule
Unpaired & &\checkmark &\checkmark & &\checkmark &\checkmark &\checkmark &\checkmark \\
Multi-domain & & &\checkmark & & & &\checkmark &\checkmark \\
MultiModal & & & &\checkmark &\checkmark &\checkmark &\checkmark &\checkmark \\
SinglePair & & &\checkmark & & & & &\checkmark \\
\bottomrule
\vspace{-0.5cm}
\end{tabular}
\end{table*}

Specifically, we use an image and its label as input to the style encoder for domain-specific representation extraction and feed the latent codes along with the target labels to the generator to generate a set of diverse target images.
The general framework of our proposed method, SoloGAN, is presented in Fig.~\ref{fig:ill_intro}(b). SoloGAN is able to learn multimodal mappings among multiple domains using a single pair of generator and discriminator. Since training samples from all domains are available to the single discriminator, its capability in discriminating real and generated samples is enhanced.

To better validate the performance of the proposed SoloGAN, we modify existing datasets to be more challenging, where the images contain complex backgrounds or the translation of these images requires shape changes.
Furthermore, we propose an evaluation method to assess the performance of I2I translation models more comprehensively.
Since the failure of I2I translation models should be taken into consideration, the Fr\'echet inception distance (FID) \cite{FID} and translation success rate are adopted.
Qualitative and quantitative comparisons against variants of SoloGAN and counterparts demonstrate the effectiveness of each component of SoloGAN and the merits of the entire method.
The main contributions of this work are:
\begin{itemize}
    \item Instead of using multiple GANs, the proposed SoloGAN adopts a single pair of generator and discriminator, where the content/style encoder and the generator are shared among multiple domains.
    \item To enhance the ability of generating diverse images of multiple target domains, the proposed SoloGAN adopts a tailored projection model discriminator with an auxiliary classifier. 
    \item We demonstrate the effectiveness of our proposed SoloGAN over a range of datasets. Across all datasets, SoloGAN yields promising results in comparison with modern counterparts, especially on some challenging datasets. For instance, SoloGAN has achieved remarkable results on the  ${cat}$$\leftrightarrow$$ {dog}$ dataset which requires extreme shape variations, and the ${horse}$$ \leftrightarrow$$ {zebra}$ dataset which retains the complex background after translation.
\end{itemize}

The remainder of the paper is organized as follows. Section~\ref{related_work} summarizes the related work. 
The proposed SoloGAN is elaborated in Section~\ref{method}. 
In Section~\ref{experiments}, we conduct the benchmark experiments together with some ablation studies to comprehensively assess the performance of SoloGAN.
Finally, we conclude with a brief summary of the proposed SoloGAN in Section~\ref{conclusion}.

\section{Related Work}\label{related_work}

\subsection{Generative Adversarial Networks}

Generative adversarial networks (GANs)~\cite{GANs} have gained much attention in recent years.
To improve the training process as well as the quality and diversity of generated samples, staple adversarial losses~\cite{lsgan, WGAN-GP} and regularization techniques~\cite{WeightN, SpectralN} have been proposed.
In addition, conditional GANs~\cite{cGAN} have been developed to help generate samples of desired classes.
For instance, the ACGAN~\cite{ACGAN}  method uses an auxiliary classifier to train the generator for generating samples of the desired classes.
I2I translation methods can also be formulated based on conditional GANs since the synthesized images belong to a specifically desired domain.

\subsection{Image-to-Image (I2I) Translation}

The I2I translation aims to learn a function to transfer the domain-specific part of a given image to the target domain~\cite{I2I}.
In the Pix2Pix~\cite{Pix2Pix} method, a paired training dataset is used to train a cGAN~\cite{cGAN} in a supervised manner.
To alleviate the issue of collecting a large amount of paired training data, the CycleGAN~\cite{CycleGAN} model uses a cycle consistency loss to preserve the key attributes between the input and the translated image.
Considering that the cycle-consistency assumption is often too restrictive, CUT~\cite{park2020contrastive} adopts contrastive learning to maintain correspondence in content by maximizing the mutual information between the corresponding input and output patches. In contrast to explicit methods, LETIT~\cite{zhao2021unpaired} applies energy-based models to perform implicit I2I translation by direct maximum likelihood estimation, having achieved impressive results with less computational cost. To further alleviate the domain labels as supervision, TUNIT~\cite{baek2021rethinking} adopts a truly unsupervised I2I translation mode. In addition, pSp~\cite{richardson2021encoding} explored StyleGAN's powerful ability to generate high-resolution images in I2I translations.
Despite these methods have demonstrated promising performance in I2I translation within a single domain, they have limited scalability in handling I2I translation among multiple domains, since different generative models should be trained for each pair of source/target domains.
%
Instead of training multiple GANs for multi-domain I2I translation, e.g. ComboGAN~\cite{ComboGAN}, a number of methods have explored multi-domain translation using a single GAN, e.g. StarGAN~\cite{StarGAN} and GANimation~\cite{GANimation}.

Numerous I2I translation problems are inherently multimodal.
The BicycleGAN~\cite{BicycleGAN} model explicitly encourages a bijection between two spaces in a supervised manner, which makes it possible to generate different images using different latent codes.
The MUNIT \cite{MUNIT} and DRIT \cite{DRIT} methods are developed based on partially shared latent space. These approaches use a content encoder as well as a style encoder to decompose the latent space of images into a domain-invariance part and a domain-specific part, respectively. 
DRIT++~\cite{lee2020drit++} introduces the mode seeking regularization proposed in ~\cite{mao2019mode} to alleviate the mode-collapse problem in DRIT. StarGANv2~\cite{choi2020stargan} also achieves promising results over face images without a complex background.
As a result, these methods can translate images while preserving the domain-invariance properties without supervision.
Closely related to this work is the recently proposed SingleGAN~\cite{SingleGAN} scheme, which shares the style encoder and generator using a conditional approach.
Nevertheless, the SingleGAN model still requires multiple discriminators to determine the domain of each image. In addition, SingleGAN does not split the latent space into domain-invariant and domain-specific parts, which results in its poor performance over datasets involving extreme shape variations.
Table~\ref{tb:comparisions} shows a comparison of existing I2I translation methods.

\section{SoloGAN}\label{method}

\begin{figure*}[t]
\centering
    \subfigure[Overall framework of SoloGAN]{
        \begin{minipage}[t]{0.59\linewidth}
            \centering
            \includegraphics[width=4.23in]{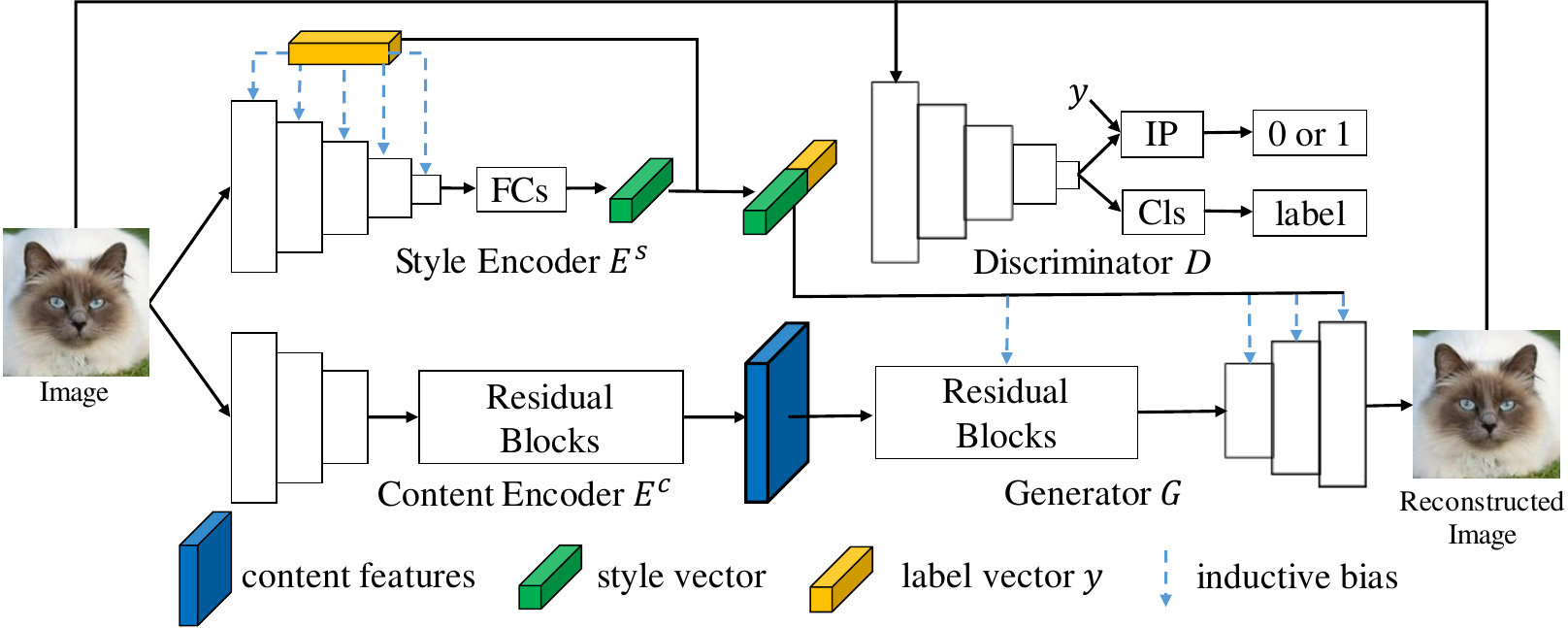}
        \end{minipage}%
    }
    \centering
    \subfigure[I2I translation condition on one-hot label vectors]{
        \begin{minipage}[t]{0.39\linewidth}
            \centering
            \includegraphics[width=2.45in]{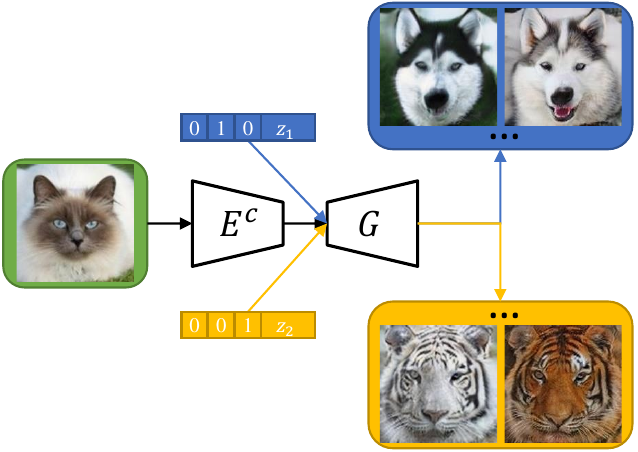}
        \end{minipage}%
    }

   \caption{SoloGAN overview. $y$ is a domain label, where label vector presented $\mathbf{y}$ is one-hot vector of $y$ (e.g. given $y = 2$, $\mathbf{y}=(0, 0, 1)$). FCs, IP, and Cls represent a few fully connected layers, inner product operation, and a classifier, respectively. Both $z$ represent style vectors randomly sampled from $N(0, 1)$, while `010' and `001' are one-hot labels for dogs and tigers, respectively.}
\label{fig:network_structure}
\vspace{-0.3cm}
\end{figure*}

The goal of this work is to learn multimodal mappings among multiple domains using a single GAN.
The overall network structure of our proposed SoloGAN is shown in Fig.~\ref{fig:network_structure}(a) and consists of an encoder $E$ (including a content $E^{c}$, style $E^{s}$), generator $G$, and discriminator $D$.
Fig.~\ref{fig:network_structure}(b) shows a sample of multimodal I2I translation among multiple domains by SoloGAN.
In addition, the Central Biasing Instance Normalization (CBIN)\footnote{
$\mathrm{CBIN}(x_i) = \frac{x_i-\mathrm{E}[x_i]}{\sqrt{\mathrm{Var}[x_i]}} +\mathrm{tanh}(f_i(z))$, where $x_i$ is the $i$-th feature map, and $f_i$ is a transformation function taking the domain label vector as input.}~\cite{CBIN} scheme is used for inductive bias.

\subsection{Encoder}
The encoder is used to map the input image into the latent space.
As mentioned above, an image can be decomposed into the content and style in the latent space. Hence, we design a content encoder and a style encoder, respectively.
The content is domain invariant and can be learned by a typical encoder. The style is encoded using a conditional encoder with the domain labels as the condition (i.e. domain labels are transferred to a one-hot vector).
Furthermore, the style is distilled in a low-dimensional vector, where the length is set to $8$ in this work.

\subsection{Generator}
Conditioned by a target domain label, the generator in SoloGAN maps the given content and style to the output directly.
Specifically, the target domain label vector is concatenated with the given style vector, and then the obtained vector is fed into the generator network by using the CBIN scheme~\cite{CBIN}. The generator learns how to maximize the probability of the discriminator distinguishing outputs from it as real. By the end, the generator can produce samples with realistic details and correct global structure.

\subsection{Discriminator}\label{discriminator}
\setlength{\tabcolsep}{1pt}
\begin{figure}[tbp]
\centering
\begin{tabular}{ccc}
        \includegraphics[width=.22\linewidth]{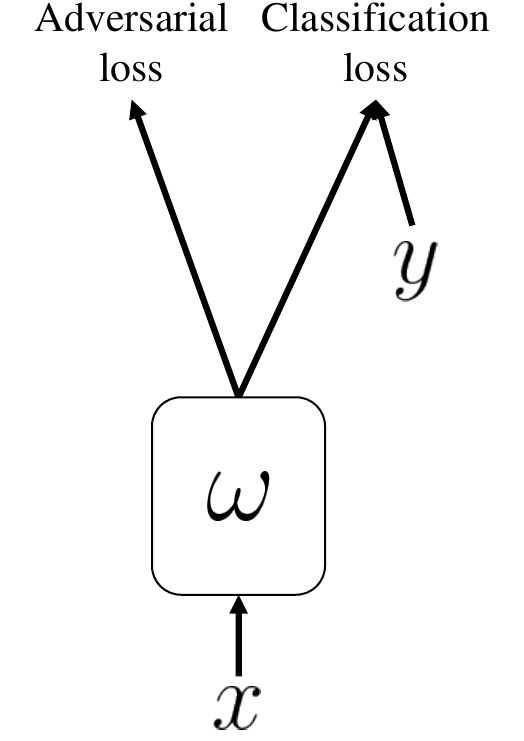} &
        \includegraphics[width=.22\linewidth]{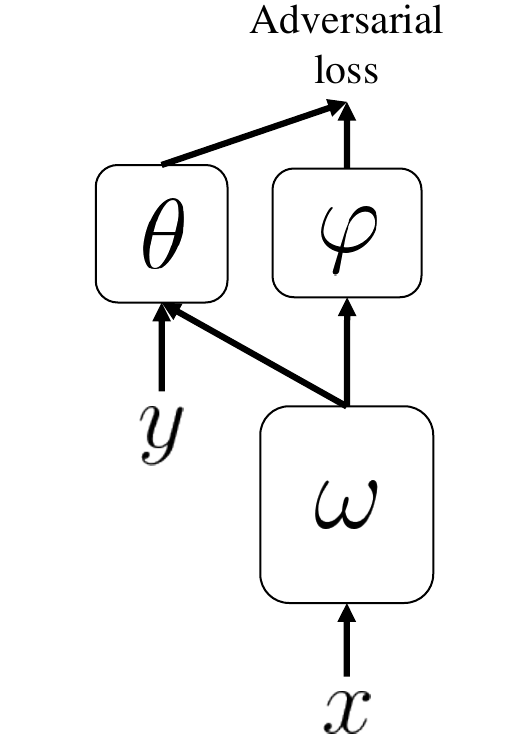} &
           \includegraphics[width=.22\linewidth]{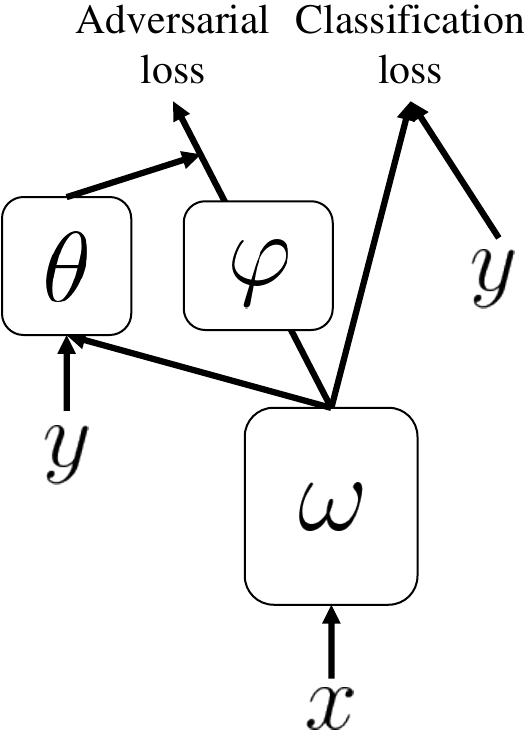}\\
           (a) Classifier & (b) Projection & (c) Projection with classifier
\end{tabular}
   \caption{Three kinds of discriminator models for conditional GANs, where $x$ and $y$ are image and label, respectively, and $\omega$, $\theta$, and $\varphi$ are learnable parameters.}
\vspace{-0.3cm}
\label{fig:discriminators}
\end{figure}

In contrast to existing I2I methods that separate multiple domains into individuals and use multiple domain-specific discriminators to distinguish the target translated images, we use a single discriminator as in conventional class-conditional image generation~\cite{ACGAN,SAGAN,BigGAN}.
Motivated by the classifier-based discriminator~\cite{ACGAN} [Fig.~\ref{fig:discriminators}(a)] and the projection discriminator (PD)~\cite{ProjectionD} [Fig.~\ref{fig:discriminators}(b)],
We propose a new discriminator [as shown in Fig.~\ref{fig:discriminators}(c)] in this work.
The classifier-based discriminator incorporates the label information into the objective function by augmenting the original discriminator objective with the likelihood score of the classifier on both the generated and training images.
On the other hand, the projection operator incorporates the label into the discriminator by taking an inner product between the embedded one-hot vector domain label $\mathbf{y}$ and the intermediate output feature vector. This significantly improves the quality of class conditional image generation.
However, the classifier-based discriminator fails to generate samples with realistic details, while the projection discriminator is ineffective in punishing the generator for translating the input image into a target domain.
In the proposed projection with a classification discriminator, part of the projection model structure is shared with an auxiliary classifier. This helps the projection discriminator classify the translated images into a target domain.
Instead of using multiscale discriminators~\cite{pixpixHD} as in most I2I GAN-based models (e.g. MUNIT~\cite{MUNIT} and DRIT~\cite{DRIT}), we use a single-scale discriminator in this work.

\subsection{Loss Functions}
  Our loss is a weighted sum of the adversarial loss, domain classification loss, cycle consistency loss, and bidirectional reconstruction loss.
These four losses aim to force the model to generate realistic samples, translate samples into the target domain, learn translation without ground truths, and decompose the content from style.

\noindent \textbf{Adversarial loss.}
To make the images generated by our generator indistinguishable from the real images in the target domains, we use the least-square adversarial loss~\cite{lsgan}:
\begin{align} \label{eq:ganloss}
&L_{adv}^{D}= \mathbb{E}_{x, y', z\sim N(0,1)}[D(\hat{x})^{2}] + \mathbb{E}_{x}[(1-D(x))^{2}], \\ 
&L_{adv}^{G} =\mathbb{E}_{x, y', z\sim N(0,1)}[(1 - D(\hat{x}))^{2}],
\end{align}
where $\hat{x}$ is an image in domain $y'$ translated from $x$ by $G(c, z, y')$, and $c$ denotes the content extracted from $x$ with $E^c$.
Given an input pair $(x, y)$ and a target domain label $y'$, the goal is to translate image $x$ from domain $y$ into domain $y'$.
The generator $G$ synthesizes an image $\hat{x}$ conditioned on both the content of $x$ (i.e. $c$) and target domain label $y'$, while the discriminator $D$ aims to classify and distinguish between real and fake.
We refer to the term $D(x)$ as the probability that an input $x$ is a real sample in the domain $y$.

\noindent \textbf{Domain classification loss}
To force the generator to translate an input image into a target domain when conditioned by a target label, we use the domain classification loss of both real and translated images when optimizing $D$, $G$, and $E$:
\begin{align}\label{eq:clsloss}
    L_{cls}^{r} = \mathbb{E}_{x, y}\left[-\log D_{cls}(y|x) \right], \\ 
    L_{cls}^{t} = \mathbb{E}_{\hat{x}, y'}\left[-\log D_{cls}(y'|\hat{x}) \right],
\end{align}
where $L_{cls}^{r}$ and $L_{cls}^{t}$ are used to optimize $D$ and the joint of $G$ and $E$, respectively.

\vspace{2mm}
\noindent \textbf{Cycle consistency loss}
When there is a lack of paired training samples for supervised learning, training $G$ with respect to the adversarial loss in (\ref{eq:ganloss}) does not guarantee that the translated images will preserve the content of the given image while only changing the style.
To alleviate this problem, we apply a cycle consistency loss \cite{CycleGAN} to the generator:
\begin{align}\label{eq:cycloss}
    L_{cyc}  = \mathbb{E}_{x, y, \hat{x}, y', s}\left[||x- G(E^c(\hat{x}), s, y)||_{1}\right],
\end{align}
where $s$ denotes the style extracted from $x$ with $E^s$.
Note that $c$, $\hat{x}$ [as given in (\ref{eq:ganloss})], and $s$ are used as consistent denotations in the following unless otherwise specified.
Specifically, an image $x$ should be reconstructed after being translated to $\hat{x}$ within the target domain $y'$.

\vspace{2mm}
\noindent \textbf{Bidirectional reconstruction loss}
To encourage a bijection between two spaces, we introduce a bidirectional reconstruction loss as proposed in the MUNIT~\cite{MUNIT} method.
When the generator maps the latent code of an image to the output, the output should be the same as the given image (image reconstruction), and an encoder should learn the mapping from the output back to the same latent code (latent reconstruction) with the following losses:
\begin{itemize}
\item \textbf{Image reconstruction.}
The image reconstruction loss requires that the translated image be reconstructed back to $x$ by recombining its content and style:
\begin{align}\label{eq:img_recloss}
    L_{rec}^{img} =  \mathbb{E}_{x, y, c, s}[||x-G(c, s, y))||_{1}].
\end{align}

\item \textbf{Latent reconstruction}
The latent reconstruction is presented in the BicycleGAN~\cite{BicycleGAN} scheme for style reconstruction to alleviate the mode collapse problem.
The MUNIT \cite{MUNIT} method adds the content reconstruction loss to encourage the preservation of the semantic content in the input image during the translation with
\begin{align}\label{eq:self_recloss}
    L_{rec}^{latent} = & \mathbb{E}_{\hat{x}, y', z\sim N(0,1)}[||z- E^s(\hat{x}, y')||_{1}] + \\ & \mathbb{E}_{\hat{x}, c}[||c - E^c(\hat{x})||_{1}].
\end{align}
\end{itemize}

\noindent \textbf{Full objective.}
We jointly train the pair of style and content encoders and generators while the discriminator is trained independently, and the final objective functions for the joint ($L_{GE}$) and discriminator ($L_D$) are:
\begin{align}\label{eq:fullObjective}
L_{GE} = & L_{adv}^{G} + \lambda_{cls}L_{cls}^{t} + \lambda_{cyc}L_{cyc} \nonumber\\
&+ \lambda_{rec}^{img}L_{rec}^{img} + \lambda_{rec}^{latent}L_{rec}^{latent} , \\
L_{D} = & L_{adv}^{D} + \lambda_{cls}L_{cls}^{r},
\end{align}
$\lambda_{cls}$, $\lambda_{cyc}$, $\lambda_{rec}^{img}$, and $\lambda_{rec}^{latent}$ are hyperparameters that control the relative importance of corresponding losses compared to the adversarial loss, which are set to 1, 10, 10, and 1, respectively, as suggested in ~\cite{MUNIT, DRIT}. 

Based on the above, the training process of SoloGAN is summarized in Algorithm~\ref{algo:framework}.

\begin{algorithm}\footnotesize
\caption{Training process of SoloGAN \label{algo:framework}}
  \KwIn{Training data $\mathcal{D}_{trn}$, max. \# of training epochs $T$, max. \# of domains $N$, Gaussian noise $P_z$. $x$, $y$ represent $N$ images and corresponding domain labels in each batch.}
  \KwOut{$G$, $E^c$, $E^s$, $D$}
  $t$ $\leftarrow$ 0 \textcolor{gray}{// initialize an epoch counter}; \\
  $G$, $E^c$, $E^s$, $D$ $\leftarrow$ Xavier \textcolor{gray}{// initialize weights with Xavier function}; \
  \While{$t < T$}
  {
    \For{$[x, y]$ $\in \mathcal{D}_{trn}$} 
    {
        \textcolor{gray}{ // Obtain target domain labels and style vectors randomly.} \\
        $y'$, $z$ $ \leftarrow N$ random numbers from [$1$, $\dots$, $N$] and samples from $P_z$ \\
        \textcolor{gray}{ // Extract the content and style feature} \\
        $c$, $s$ $\leftarrow$ $E^c$($x$),  $E^s$($x$, $y$) \\
        \textcolor{gray}{ // Generate fake images } \\
        $\hat{x}$ $\leftarrow$ $G$($c$, $z$, $y'$) \\
        \textcolor{black}{/ ********* Update the discriminator ******** /} \\
        Update $D$ according to Eq.~10 \\
        \textcolor{black}{/ ********* Update generator and encoder ******** /} \\
        \textcolor{gray}{ // Reconstruction of images, content and style features} \\
        $x_{rec}$, $c_{rec}$, $z_{rec}$ $\leftarrow $ $G$($c$, $s$, $y$), $E^c$($\hat{x}$), $E^s$($\hat{x}$, $y'$)  \\
        Update $G$, $E^c$, and $E^s$ according to Eq.~9. \\
        }
  $t$ $\leftarrow$ $t + 1$ \textcolor{gray}{// repeat above steps for $T$ epochs}.
 }
 \Return $G$, $E^c$, $E^s$, $D$\
\end{algorithm}

\begin{table}[tbp]
\centering
\scriptsize
\caption{Structures of proposed style and content encoder.}
\label{tb:style_content}
\begin{tabular}{c|c}
\toprule
\multicolumn{2}{c}{Input RBG image $x$ $\in \mathbb{R}^{256\times256\times3}$}                                            \\ \hline
CONV-(C64, K4x4, S2, P1)                & \begin{tabular}[c]{@{}c@{}}CONV-(C64, K7x7, S1, P3), IN, ReLU\end{tabular}     \\ \hline
CD-ResBlock-(C128)                      & \begin{tabular}[c]{@{}c@{}}CONV-(C128, K4x4, S2, P1), IN, ReLU\end{tabular} \\ \hline
CD-ResBlock-(C256)                      & \begin{tabular}[c]{@{}c@{}}CONV-(C256, K4x4, S2, P1), IN, ReLU\end{tabular} \\ \hline
CD-ResBlock-(C256)                      & R-ResBlock-(C256)                                                                 \\ \hline
CBIN, ReLU                              & R-ResBlock-(C256)                                                                 \\ \hline
GAP                                     & R-ResBlock-(C256)                                                                 \\ \hline
FC-(8)                                  & R-ResBlock-(C256)                                                                 \\ \hline
Output $s \in \mathbb{R}^{8}$           & Output $c \in \mathbb{R}^{64\times64\times256}$                                   \\ \hline
\textbf{Style Encoder}                  & \textbf{Content Encoder}
\\
\bottomrule
\end{tabular}
\end{table}

\begin{table}[tbp]
\centering
\caption{Detailed structures of our proposed generator and discriminator.
$\mathbf{h}$ is output of global average pooling layer (GAP), $d$ is result from fully connected layer given $\mathbf{h} $ as input, and n denotes number of domains.}
\label{tb:generator_discriminator}
\scriptsize
\begin{tabular}{c|c}
\toprule
Input content $c$ $\in \mathbb{R}^{64\times64\times256}$, style $s \in \mathbb{R}^{8}$ & Input RBG image $x$ $\in \mathbb{R}^{256\times256\times3}$                                                   \\ \hline
CONV-(C256, K3x3, S1, P1), CBIN, ReLU                    & CONV-(C64, K4x4, S2, P1), LReLU                                                                              \\ \hline
C-ResBlock-(C256)                                        & CONV-(C128, K4x4, S2, P1), LReLU                                                                             \\ \hline
C-ResBlock-(C256)                                        & CONV-(C256, K4x4, S2, P1), LReLU                                                                             \\ \hline
C-ResBlock-(C256)                                        & CONV-(C512, K4x4, S2, P1), LReLU                                                                             \\ \hline
C-ResBlock-(C256)                                        & GAP                                                                                                          \\ \hline
C-ResBlock-(C256), CBIN, ReLU                            & FC-(1)                                                                                                       \\ \hline
TrCONV-(C128, K4x4, S2, P1), CBIN, ReLU                  & $dis$ = Embed($y$)$ \cdot \mathbf{h} + d$                                                                        \\ \hline
TrCONV-(C64, K4x4, S2, P1), CBIN, ReLU                   & CONV-(C1024, K4x4, S2, P1), LReLU                                                                   \\ \hline
CONV-(C3, K7x7, S1, P3), Tanh                            & $cls$ = CONV-(Cn, K4x4, S1, P0)                                                                              \\ \hline
Output RBG image $\hat{x} \in \mathbb{R}^{256\times256\times3}$    & Output $dis \in \mathbb{R}^{1}$, $cls \in \mathbb{R}^{\text{n}} $ \\
\textbf{Generator}                  & \textbf{Discriminator}
\\
\bottomrule
\end{tabular}
\end{table}

\begin{table}[t!]
\centering
\caption{Statistics of evaluated datasets.}
\label{tb:datasets}
\scriptsize
\begin{tabular}{c|cccccccccccc}
\toprule
\multicolumn{1}{c|}{Dataset}
& \multicolumn{6}{c}{${cat} \leftrightarrow {dog} \leftrightarrow {tiger}$}                                                                                               & \multicolumn{6}{c}{${black} \leftrightarrow {blond} \leftrightarrow {brown}$}                                                                                           \\
\cline{3-6}
\cline{9-12}

\multicolumn{1}{c|}{\multirow{3}{*}{Samples}}
& \multicolumn{2}{c}{cat}   & \multicolumn{2}{c}{dog}    & \multicolumn{2}{c}{tiger}
& \multicolumn{2}{c}{black} & \multicolumn{2}{c}{blond}  & \multicolumn{2}{c}{brown}
\\

& \multicolumn{1}{c}{train} & \multicolumn{1}{c}{test} & \multicolumn{1}{c}{train} & \multicolumn{1}{c}{test}
& \multicolumn{1}{c}{train} & \multicolumn{1}{c}{test} & \multicolumn{1}{c}{train} & \multicolumn{1}{c}{test}
& \multicolumn{1}{c}{train} & \multicolumn{1}{c}{test} & \multicolumn{1}{c}{train} & \multicolumn{1}{c}{test}
\\

\multicolumn{1}{c}{}
& \multicolumn{1}{|c}{771}  & \multicolumn{1}{c}{100}  & \multicolumn{1}{c}{1264}  & \multicolumn{1}{c}{100}
& \multicolumn{1}{c}{1173}  & \multicolumn{1}{c}{100}  & \multicolumn{1}{c}{5000}  & \multicolumn{1}{c}{300}
& \multicolumn{1}{c}{5000}  & \multicolumn{1}{c}{300}  & \multicolumn{1}{c}{5000}  & \multicolumn{1}{c}{300}
\\
\hline
\multirow{2}{*}{Dataset}
& \multicolumn{4}{c}{${summer} \leftrightarrow {winter}$}
& \multicolumn{8}{c}{${edges} \leftrightarrow {bags\&shoes}$}                                                                                                                    \\
\cline{3-4}
\cline{7-12}
& \multicolumn{2}{c}{summer}  & \multicolumn{2}{c}{winter}
& \multicolumn{2}{c}{edges}   & \multicolumn{2}{c}{bags}
& \multicolumn{2}{c}{edges}   & \multicolumn{2}{c}{shoes}         \\
\multirow{2}{*}{Samples}
& train  & test  & train  & test  & train  & test  & train  & test  & train  & test  & train   & test  \\
& 1231   & 309   & 962    & 238   & 2500    & 150   & 2500   & 150   & 2500  & 150  & 2500   & 150   \\
\hline

\multirow{2}{*}{Dataset}
& \multicolumn{4}{c}{${horse} \leftrightarrow {zebra}$}                                                           & \multicolumn{8}{c}{${leopard} \leftrightarrow {lion} \leftrightarrow {tiger} \leftrightarrow {bobcat}$}                                                                                                                          \\
\cline{3-4}
\cline{7-12}
& \multicolumn{2}{c}{horse}    & \multicolumn{2}{c}{zebra}    & \multicolumn{2}{c}{leopard}                          & \multicolumn{2}{c}{lion}     & \multicolumn{2}{c}{tiger}    & \multicolumn{2}{c}{bobcat}                           \\
\multirow{2}{*}{Samples}
& train  & test  & train  & test  & train  & test  & train  & test  & train  & test  & train  & test   \\
& 1545   & 100   & 1070   & 100   & 620    & 100   & 919    & 100   & 777    & 100   & 530    & 100     \\
\hline

\multirow{2}{*}{Dataset}
& \multicolumn{4}{c}{${day} \leftrightarrow {night}$}                                                           & \multicolumn{8}{c}{${photo}\leftrightarrow {Monet} \leftrightarrow {VanGogh} \leftrightarrow {Cezzan}$}                                                                                                                         \\
\cline{3-4}
\cline{7-12}
& \multicolumn{2}{c}{day}           & \multicolumn{2}{c}{night}        & \multicolumn{2}{c}{photo}                           & \multicolumn{2}{c}{Van Gogh}      & \multicolumn{2}{c}{Monet}        & \multicolumn{2}{c}{Cezzan}                           \\
\multirow{2}{*}{Samples}
& train  & test  & train  & test   & train  & test  & train  & test  & train  & test  & train  & test    \\
& 1000   & 100   & 993    & 100    & 1231   & 309   & 400    & 400   & 1072   & 121   & 525    & 58     \\
\bottomrule
\end{tabular}
\vspace{-0.3cm}
\end{table}
\section{Experiments}   \label{experiments}
We describe implementation details  (including network and training details) and then introduce evaluation metrics as well as datasets in the experiments.
The effectiveness of our model is validated by ablation studies and comparisons with other state-of-the-art I2I translation methods.
We then discuss the limitations of the SoloGAN method.

\subsection{ Network Details}

\begin{figure}[tbp]
\begin{center}
\begin{tabular}{ccc}
\includegraphics[width=1.242in]{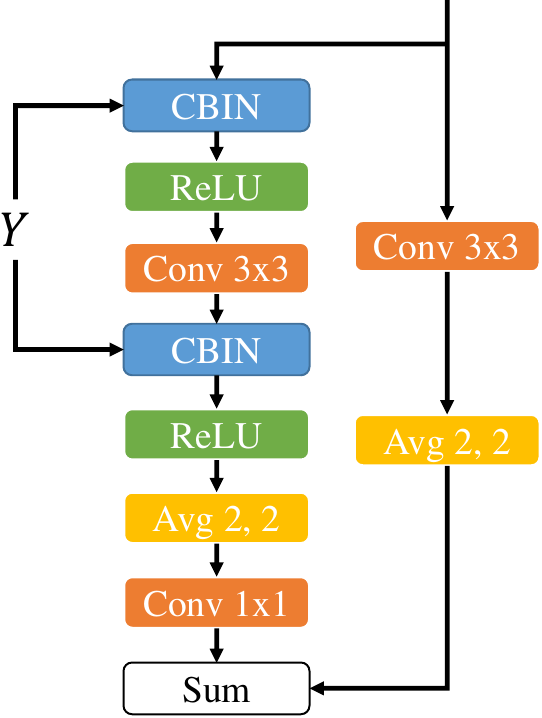} &
\includegraphics[width=0.588in]{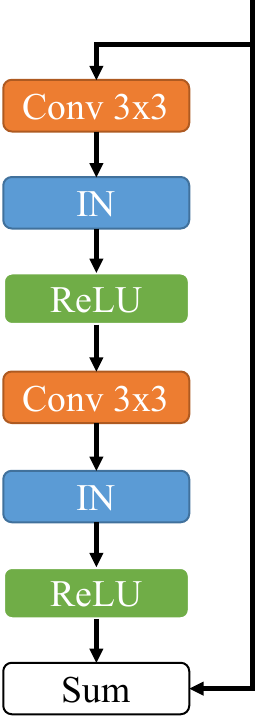} &
\includegraphics[width=1.5in]{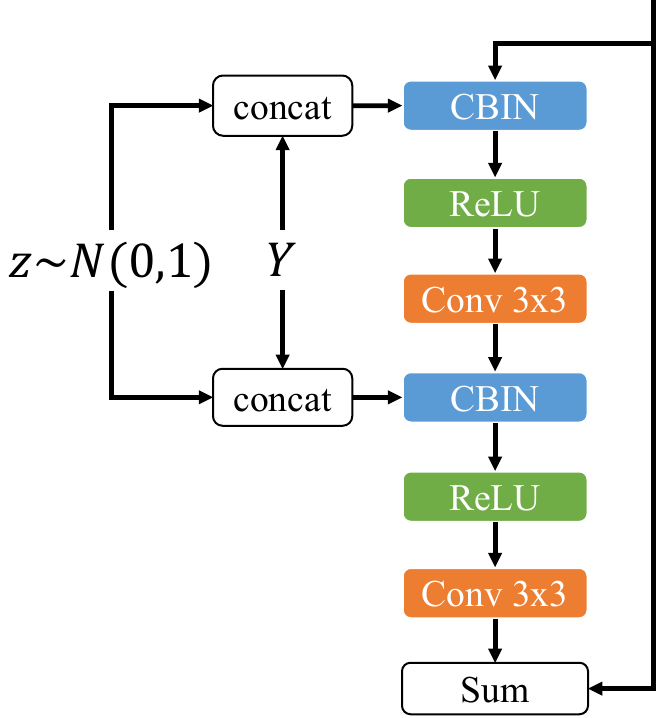} \\
                          (a) & (b) & (c)
\end{tabular}
\end{center}
\vspace{-0.2cm}
\caption{(a) Conditional downsample residual block (CD-ResBlock) in our proposed style encoder, where ``Avg 2, 2'' denotes average pooling layer with both kernel and stride size set to 2.
(b) Regular residual block (R-ResBlock) in our proposed content encoder.
(c) Conditional residual block (C-ResBlock) used in our proposed generator.}
\label{fig:blocks_structure}
\vspace{-0.4cm}
\end{figure}

The network structures of the style encoder and content encoder are given in Table~\ref{tb:style_content}, and the structures of the generator and discriminator are given in Table~\ref{tb:generator_discriminator}.
In the tables, C, K, S, and P are the number of output channels, kernel size, stride size, and padding size, respectively. In addition, we denote the instance normalization, global average pooling, leaky ReLU, and transposed CONVolution layers with IN, GAP, LReLU, and TrCONV, respectively.
Moreover, the architectures of three different ResBlocks are shown in Fig.~\ref{fig:blocks_structure}.

\subsection{  Training Details}
  
In this study, the spectral normalization (SN)~\cite{SpectralN} method is applied to the weights of the discriminator, generator, and encoder in the training process,
where the spectral norm of each layer is restricted.
For all of the experiments, the input image is $256\times 256$ pixels, and the Adam optimizer~\cite{Adam} with $\beta_{1}=0.5, \beta_{2}=0.999$ is used to train our model.
Each minibatch consists of one image from each domain.
Xavier initialization is used to assign the initial network weights of $E$, $G$, and $D$.
The initial learning rate of $E$, $G$, and $D$ is 0.0002 for the first $n$ epochs. The initial learning rate decays to zero linearly in the remaining $n$ epochs, where $n$ is set to 50 unless otherwise noted.
The source code and trained models will be available to the public.

\setlength{\tabcolsep}{3pt}

\subsection{Datasets}
{\color{black}Table~\ref{tb:datasets} shows the datasets and statistics in the experiments. The properties of each dataset are given as follows.}
\begin{compactitem}
\item ${Day} \leftrightarrow {night} $. The images are obtained from the Transient Attributes dataset~\cite{trans_attr} with different cloud patterns and lighting conditions.

\item ${Summer} \leftrightarrow {winter}$. This dataset is used for translation of landscapes in summer and winter~\cite{CycleGAN}.

\item ${Edges}\leftrightarrow {bags\&shoes}$. This dataset is used to translate images between edges and real content (i.e. handbags and shoes).
A set of images are randomly sampled with equal probability from the edges2hangbags and edges2shoes sets~\cite{Pix2Pix}, which contain thousands of images of shoes and handbags.

\item ${Horse} \leftrightarrow {zebra}$ and ${leopard} \leftrightarrow {lion} \leftrightarrow {tiger} \leftrightarrow {bobcat}$. The images are obtained from the Animals With Attributes dataset~\cite{Animal_AWA}. These images contain objects at different scales across different backgrounds.

\item ${Cat} \leftrightarrow {dog} \leftrightarrow {tiger}$. The dog and cat images are from ~\cite{DRIT}, and we collected the tiger pictures. The translation dataset among these domains needs to account for large shape changes.

\item  ${Black} \leftrightarrow {blond} \leftrightarrow {brown} $.
This dataset is used to translate the color of human hair, which is randomly selected from the CeleA~\cite{CelebA} dataset containing 202,599 face images with 40 attributes.

\item ${Photo}\leftrightarrow {VanGogh} \leftrightarrow {Monet} \leftrightarrow {Cezanne}$.
We extract images of the VanGogh, Monet, and Cezanne paintings from the photo2vangogh, photo2monet, and photo2cezanne datasets~\cite{CycleGAN}.
  In addition, the ``photo" shows the summer images from the {summe2winter} dataset~\cite{CycleGAN}.  
\end{compactitem}

\subsection{Quantitative Evaluation Metrics}
 In this section, we present the metrics for quantitatively evaluating the translated images.

\vspace{2mm}
\noindent \textbf{Quality.}
The inception score (IS)~\cite{ImprovedGAN} and Fr\'echet inception distance (FID)~\cite{FID} are two widely used metrics for measuring the quality of the images generated by GANs. For IS, we use an Inception-V3 classifier~\cite{InceptionV3} fine-tuned on our specific dataset, and 10k translated images (100 input images and 100 translated samples per given input) are used to evaluate the IS. For FID, we use the ImageNet-pretrained Inception V3~\cite{InceptionV3} with 100 input images from each domain and 10 translated samples per given input. A lower FID value indicates a higher quality image, whereas a higher IS score indicates a better result.

\begin{figure*}[tp]
\centering
    \includegraphics[width=0.90\textwidth]{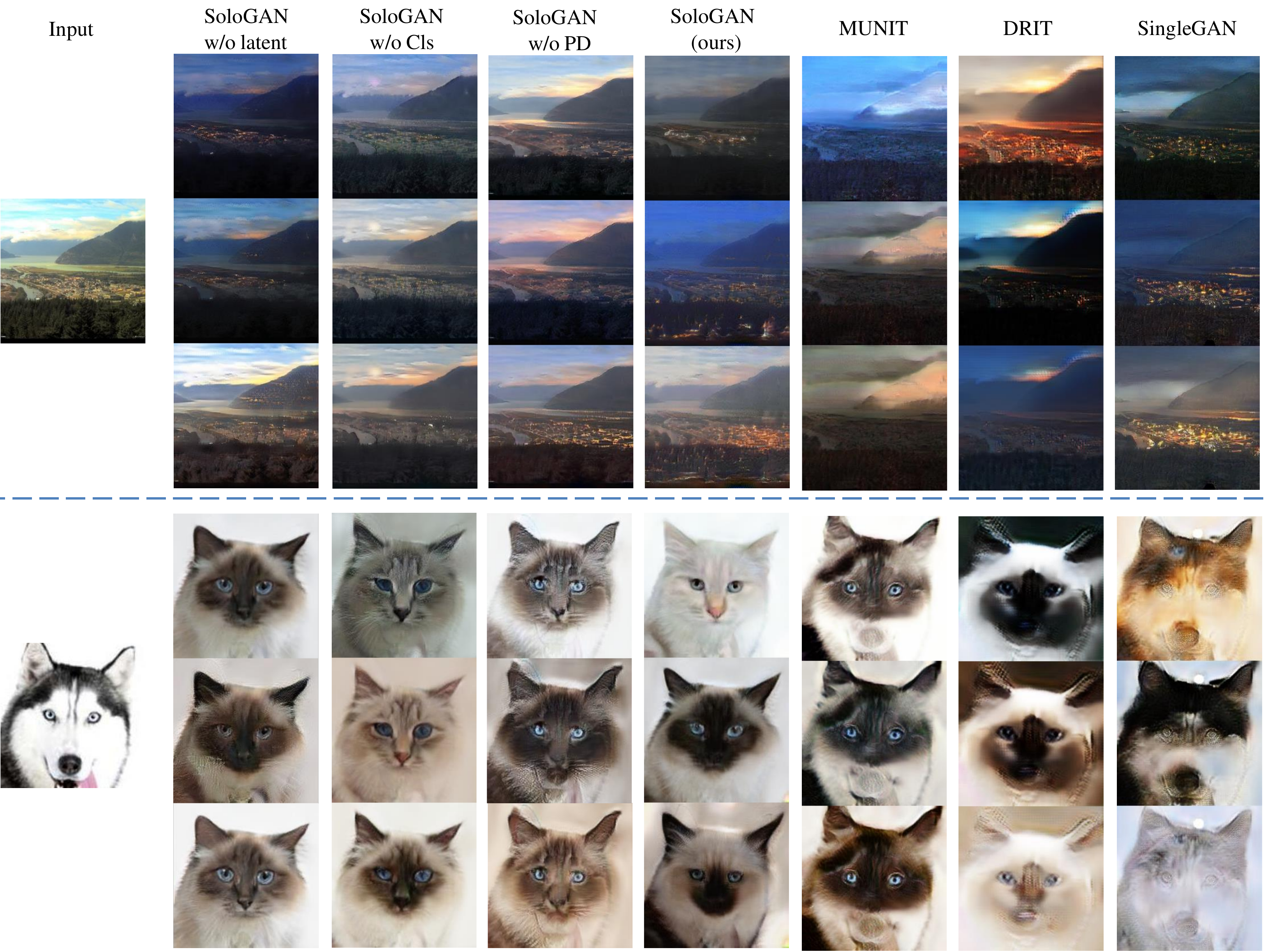}
    \caption{Qualitative comparison between SoloGANs and existing methods on ${day} \rightarrow {night}$ (top) and ${dog} \rightarrow {cat}$ (bottom) datasets. In each dataset, first column shows input, and each subsequent column shows 3 random outputs from a method.}
\label{fig:qualitative_results}
\vspace{-0.3cm}
\end{figure*}

\vspace{2mm}
\noindent \textbf{Diversity.}
For diversity assessment, we introduce the LPIPS distance~\cite{LPIPS}, which is computed by the weighted $L_{2}$ distance between deep features of 19 paired images per given input. There are 100 different input images per domain, and ImageNet-pretrained AlexNet~\cite{AlexNet} is used as the feature extractor. In addition, we introduce the conditional inception score (CIS)~\cite{MUNIT}, which is modified from IS to measure the diversity of outputs conditioned on a single input image. It is calculated as the IS but with a different estimated equation as given in~\cite{MUNIT}. A higher value of either IS or CIS indicates an image with better diversity.

\vspace{2mm}
\noindent \textbf{Translation success rate.}
Since the above evaluation metrics are not effective in explicitly reflecting whether a translation is successful, we propose to compute the translation success rate (TSR) of translated images as an additional performance metric. Specifically, we use the abovementioned fine-tuned Inception-V3 model, with a classification error of 0.50\% on the test set, as a classifier. We then perform I2I translation on 100 test images for each domain, where 10 translated images given each test image are selected to make a new test set. Finally, we classify those images with annotated target domain labels. A lower classification error ($Cls\_{err}$) indicates that the model has a higher success rate to translate input images into target domains, and $TSR = 1 - Cls\_{err}$. 

\vspace{2mm}
\noindent\textbf{Human evaluation.}
Human evaluation is much more persuasive than objective metrics for I2I translation. We conduct a user study to evaluate the realism of the translated images. We study $2\times n$ input images for each translation between two domains, e.g. $n$ images of the cat are input for translation from the cat to the dog, while $n$ images of the dog need to be translated to cats. In this work, we set $n$ to 50 and choose ${cat} \leftrightarrow {dog}$ as well as ${horse} \leftrightarrow {zebra}$. We use these two representative translation datasets which need to account for large shape changes or maintain the complex background unchanged.

\subsection{Evaluated Methods}
Three variants of SoloGAN, i.e. w/o latent, w/o Cls, and w/o PD, are presented to validate the functionality of each component in SoloGAN. The first two variants ablate $L_{rec}^{latent}$ and the classifier, and the third variant replaces the projection model discriminator with a conventional discriminator.

We evaluate the proposed method against MUNIT~\cite{MUNIT}, DRITs~\cite{DRIT, lee2020drit++}, SingleGAN~\cite{SingleGAN} and StarGANs~\cite{StarGAN, choi2020stargan} models. The MUNIT and DRITs models focus on unpaired multimodal I2I translation via disentangled representations. Among them, DRIT++~\cite{lee2020drit++} is an improved version of DRIT~\cite{DRIT} in terms of diversity. SingleGAN shares the generator using the domain labels as a condition, while multiple discriminators are required for distinguishing images in different domains. The StarGANs aim to address the issue of using multiple GANs for translation among multiple domains. StarGANv1~\cite{StarGAN} and v2~\cite{choi2020stargan} are unimodal and multimodal methods, respectively. Given a number of $n$ domains, MUNIT and DRITs need to train $n\times(n - 1)$ pairs of generators and discriminators, SingleGAN trains a generator but with $n$ discriminators, while \emph{StarGANs and our SoloGAN only train a single pair of generator and discriminator}.

\subsection{Empirical Results}
\label{sec:results}
We first provide an ablation study to analyse each component in our SoloGAN. Then, we discuss the multimodal I2I translation results over four two-domain datasets achieved by SoloGAN, MUNIT, DRITs, SingleGAN, and StarGANv2. Finally, we compare SoloGAN with StarGAN over three- and four-domain datasets. For fair comparison, we train all the evaluated methods for 100 epochs unless otherwise specified. Notably, the ${cat}$ $\leftarrow $${dog}$ dataset is derived from the ${cat} \leftrightarrow {dog} \leftrightarrow {tiger}$ database.

\noindent{  \textbf{Ablation study}. It can be observed from Fig.~\ref{fig:qualitative_results} that the SoloGAN method suffers from a serious mode collapse problem without $L_{rec}^{latent}$ (w/o latent, Column$2$). To be more specific, the SoloGAN is less effective in generating diverse and realistic images without the classifier (w/o Cls, Column$3$) or projection discriminator (w/o PD, Column$4$). There are similar phenomena in the quantitative results from Table~\ref{tb:quantitative_results}. Without $L_{rec}^{latent}$ (Line 2), the LPIPS score of SoloGAN drops dramatically from 0.300 to 0.217. When using the projection (Line 3) or the classifier (Line 4) only in our discriminator, the SoloGAN method does not perform well in terms of the LPIPS, IS, or FID values. This indicates that the entire proposed projection discriminator is important for I2I translation among multiple domains. Furthermore, the translation failure rate ($1 - TSR$) achieved by these two variants are approximately 10 times as large as those generated by SoloGAN. These results indicate that each proposed component plays an important role in translating images into target domains.
}

\begin{table*}[!htbp]
\centering
\caption{Quantitative results, where `G', `D', `$n$', `Params.', and `Time' represent generator, discriminator, the number of domains, parameters of the generator (including encoder) for $n$ domains, and time consumption when translating an input image to 100 images of a target domain, respectively. Besides, $\dagger$ denotes models require multiple discriminators. The best result in each column is highlighted in \textbf{bold}.}
\label{tb:quantitative_results}
\begin{threeparttable}
\begin{tabular}{l|c|c|c|c|c|c|c}
\toprule
\multirow{2}{*}{Method} & $day \leftrightarrow night$ & \multicolumn{4}{c|}{$ cat \leftrightarrow dog $} & \multirow{2}{*}{\# Params. $\downarrow$} & \multirow{2}{*}{\# Time   $\downarrow$} \\ \cline{2-6}
            & LPIPS $\uparrow$    & TSR $\uparrow$    & CIS $\uparrow$    & IS $\uparrow$       & FID $\downarrow$   &   & \\
\hline
\footnotesize \emph{Variants of SoloGAN}: \\ 
SoloGAN                 & 0.300       & 99.95 \% & 1.031     & 1.050    & 0.226     & 13.99 M &1.053 s \\
SoloGAN w/o latent      & 0.217       & 99.95 \% & 1.045     & 1.018    & 0.221  &13.99 M & 1.053 s \\
SoloGAN w/o Cls         & 0.253       & 99.60 \%        & 1.028     & 1.062    & 0.244   & 13.99 M   & 1.053 s \\
SoloGAN w/o PD          & 0.254       & 99.50 \%        & 1.074     & 1.030    & 0.304   & 13.99 M   & 1.053 s \\
\hline
\footnotesize \emph{Multiple pairs of G\&D}: \\
MUNIT~\cite{MUNIT}      & 0.191       & 94.60 \%        & 1.075     & 1.185    & 0.490     &$n\times (n-1) \times 15.03$ M$^{\dagger}$ & 1.447 s \\
DRIT~\cite{DRIT}        & 0.280      & 94.90 \%        & 1.048     & 1.172     & 0.753     &$n\times (n-1) \times 10.65$ M$^{\dagger}$ & 0.976 s\\
DRIT++~\cite{lee2020drit++} &   0.304    & 96.65 \%  & 1.082 &  1.102 &  0.475       &$n\times (n-1) \times 10.65$ M$^{\dagger}$ & 0.976 s \\ SingleGAN~\cite{SingleGAN}  & \textbf{0.323} & 67.85 \%   & \textbf{1.113}  & \textbf{1.315}  & 0.732 & \textbf{9.8} M$^{\dagger}$ &\textbf{0.795} s\\
\hline
\footnotesize \emph{Single pair of G\&D}: \\ 
StarGANv2~\cite{choi2020stargan} &  0.311   & \textbf{99.96} \%       & 1.021 & 1.025    & \textbf{0.186}  & 57.25 M & 1.218 s \\ 
\bottomrule
\end{tabular}
\end{threeparttable}
\vspace{-0.2cm}
\end{table*}

\begin{figure}[!htbp]
\centering
    \includegraphics[width=0.46\textwidth]{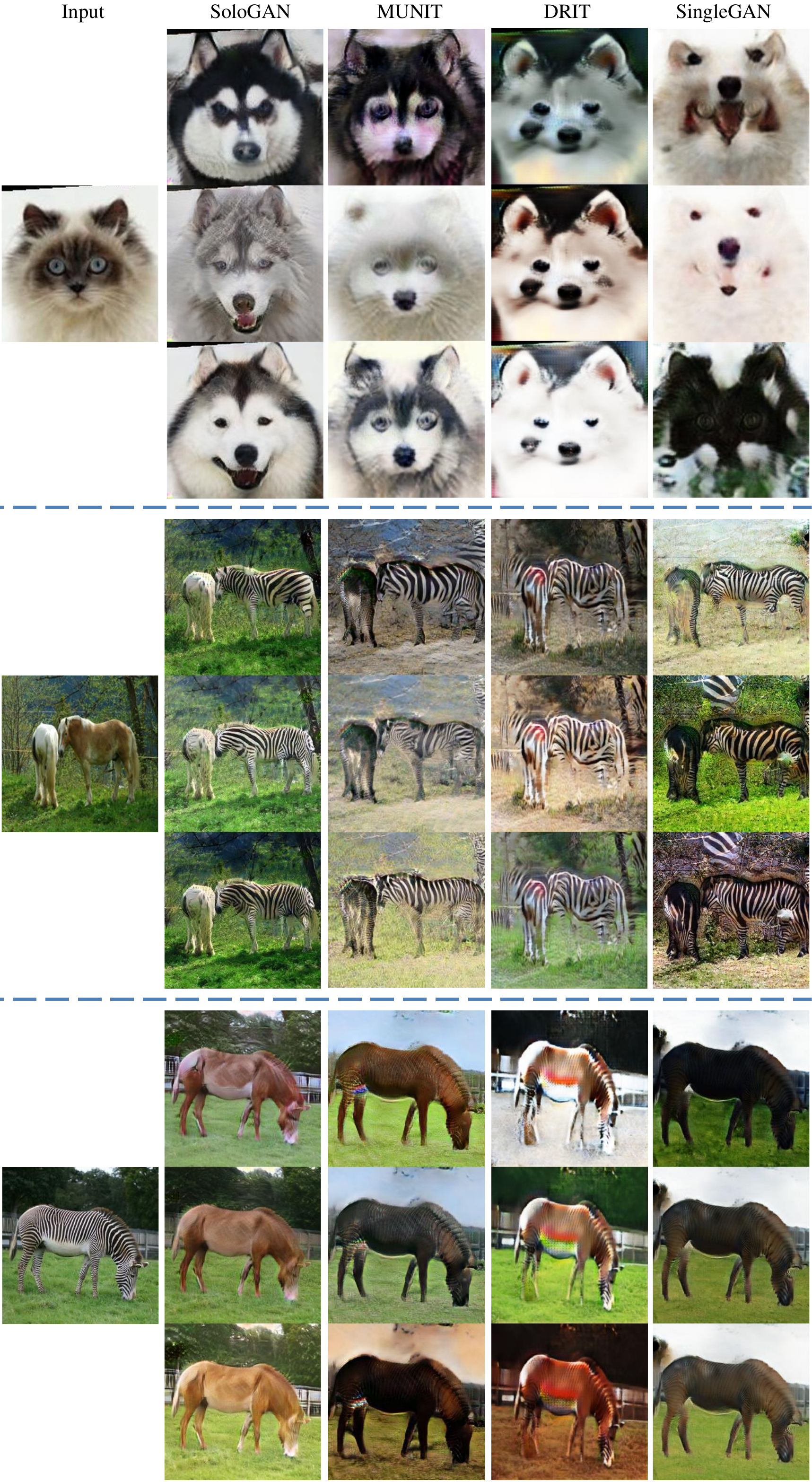}
    \caption{Additional qualitative comparison between SoloGAN, MUNIT, DRIT, and SingleGAN on ${cat} \rightarrow {dog}$ (top), ${horse} \rightarrow {zebra}$ (middle), and ${zebra} \rightarrow {horse}$ (bottom) datasets. In each dataset, first column shows input, and each subsequent column shows 3 random outputs from a method.}
\label{fig:additional_comparison_results}
\vspace{-0.45cm}
\end{figure}

\noindent{  \textbf{Multimodal I2I translation}. 
Compared with the proposed SoloGAN (Fig.~\ref{fig:qualitative_results} Column 5), three state-of-the-art multimodal frameworks with multiple generators or discriminators (i.e., MUNIT, DIRTs, and SingleGAN) show unsatisfactory performance over I2I translation datasets involving significant shape changes, e.g., translating a dog to cat images (the rightmost three columns in Fig.~\ref{fig:qualitative_results}). Meanwhile, the three compared methods (especially SingleGAN) are less effective in I2I translation on the ${horse}$$\leftrightarrow$${zebra}$ dataset (Fig.~\ref{fig:additional_comparison_results}, rightmost three columns). Although those methods can translate horse into zebra images and vice versa, they cannot effectively retain complex backgrounds, as the color of the grassland is changed for the translation from zebra to horse. This phenomenon occurs because the color of grassland in most training zebra images is not green. Therefore, independent training horse content encoders discard the green color that they believe is not the domain-invariance part, but it truly is. This can also explain why the trees behind the zebra are removed after translation into horses. Generally, these existing methods fail to translate images involving \emph{extreme shape variations} and \emph{complex backgrounds}, while our SoloGAN can perform well in those challenging datasets.

These visual observations are confirmed when evaluated with the seven objective metrics provided in Table~\ref{tb:quantitative_results}. As evidenced by Table~\ref{tb:quantitative_results}, DRIT++ indeed generates more diverse output than DRIT, and the image quality is also improved. The proposed SoloGAN model performs well in generating diverse images on the ${day}$$\leftrightarrow$${night}$ dataset and is comparable to the SingleGAN while outperforming the MUNIT and DRIT schemes (Column 2, ``LPIPS score"). In terms of the translation success rate, SoloGAN performs favorably against the evaluated schemes; by contrast, given the translation success rate of 67.85\% and some complete translation failures as in Fig.~\ref{fig:qualitative_results}, SingleGAN is unable to deal with the datasets such as ${cat}$$\leftrightarrow$${dog}$, despite the highest IS and CIS scores as achieved. In contrast to the IS and CIS metrics, which does not take any reference during its evaluation, the FID score is designed to compare the statistics of synthetic samples and real-world samples. For instance, it takes real dogs as references when evaluating the generated dogs, and its score is high when the generated samples of expected dogs are cats. This characteristic enables the FID to detect translation failures. Therefore, the FID score can reflect how realistic and diverse the generated samples are and whether the translation fails. The FID value achieved by our model is 0.226, which is much lower than that achieved by any other state-of-the-art method. This indicates the effectiveness of the SoloGAN in handling datasets involving \emph{extreme shape variations}. We also investigate the \emph{efficiency} of all the translation methods in terms of parameters of the generator (including encoders) and the latency by using a single Tesla V100. Different from the MUNIT and DRITs methods, the number of parameters for generators in our model does not increase when the number of domains increases. It is worth noting that, despite that SingleGAN has a smaller generator than SoloGAN, it requires multiple discriminators during training.
}

\begin{figure}[!htbp]
\begin{center}
    \includegraphics[width=0.43\textwidth]{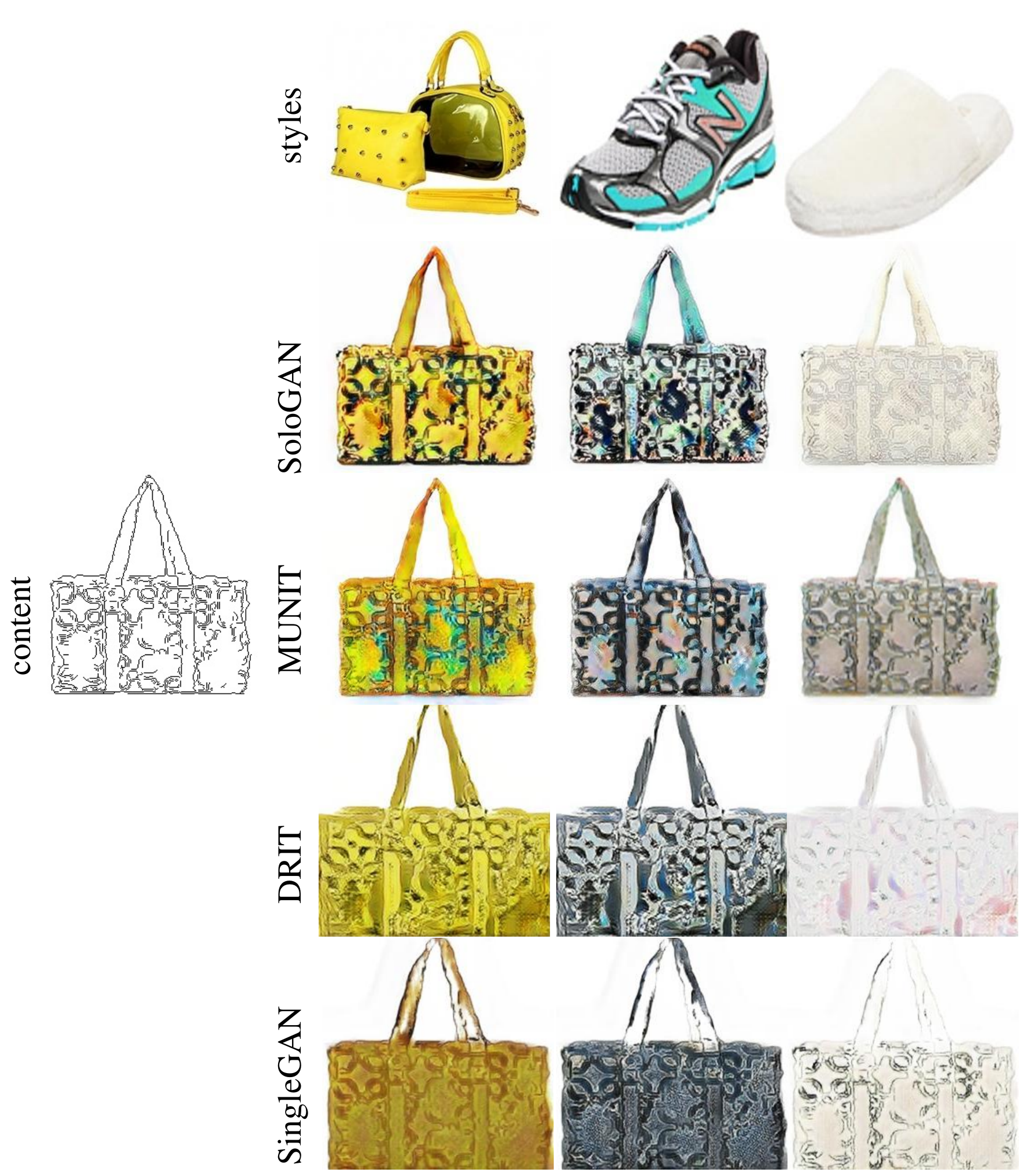}
   \caption{Example-guided I2I translations on ${edges}$$ \rightarrow$$ {bags \text{ / } shoes}$ dataset. First row and column are given three style images and one content image, respectively. Starting from second row and column, each row shows example-guided outputs from a method.}
\label{fig:example_guide}
\end{center}
\vspace{-0.4cm}
\end{figure}

Additionally, we conduct an experiment to study the disentangled representation of content and style vectors from SoloGAN and other compared frameworks by the task of example-guided translation. When given a content image $x_1$ from domain $y_1$ and a style image $x_2$ from domain $y_2$, the model generates an image $\hat{x_1}$ that recombines the content of $x_1$ and the style of $x_2$ by $G(E^c(x_1), E^s(x_2), y_2)$. Some example-guided translation results are shown in Fig.~\ref{fig:example_guide}. It can be observed that the styles of outputs from our method are more similar to the given styles than other methods, demonstrating its success in disentangling contents and styles with the same content encoder and style encoder. 

\begin{figure}[t]
\vspace{-0.2cm}
\begin{center}
    \includegraphics[width=0.41\textwidth]{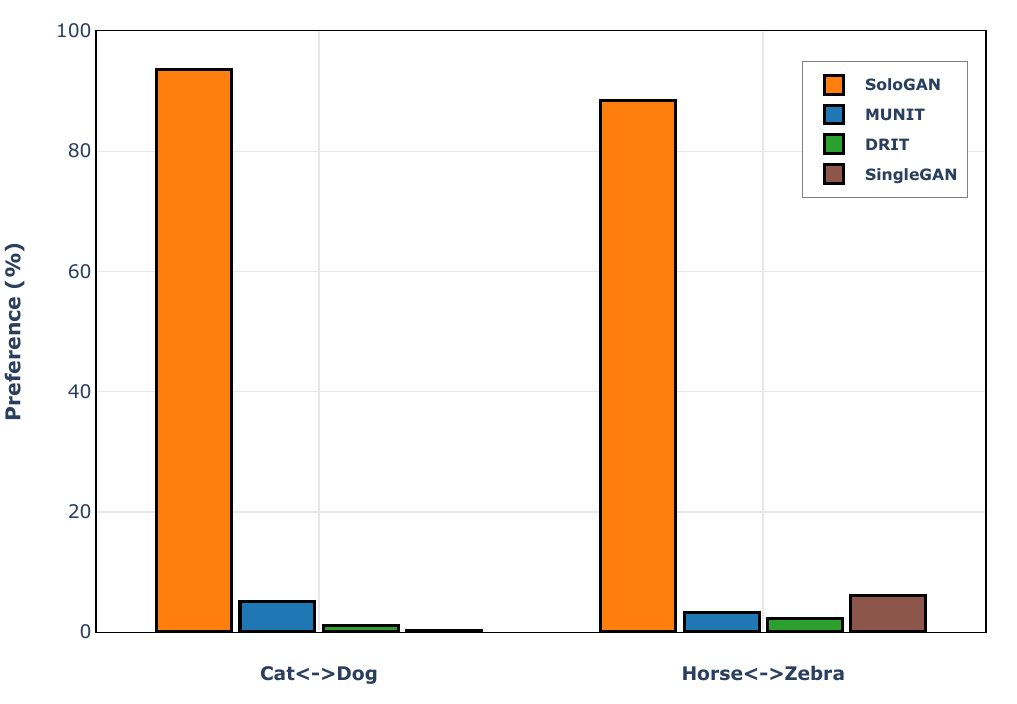}
   \caption{Human evaluation results collected from 100 reports, where vertical axis indicates percentage of preference. Higher value indicates a majority of users prefer images generated by corresponding method in terms of realistic details and translation success. }
\label{fig:user_study}
\end{center}
\vspace{-0.53cm}
\end{figure}

Fig.~\ref{fig:user_study} shows the user study results. The images synthesized by SoloGAN are considered more realistic than those synthesized by the other schemes.

\begin{figure}[t]
\begin{center}
    \includegraphics[width=0.48\textwidth]{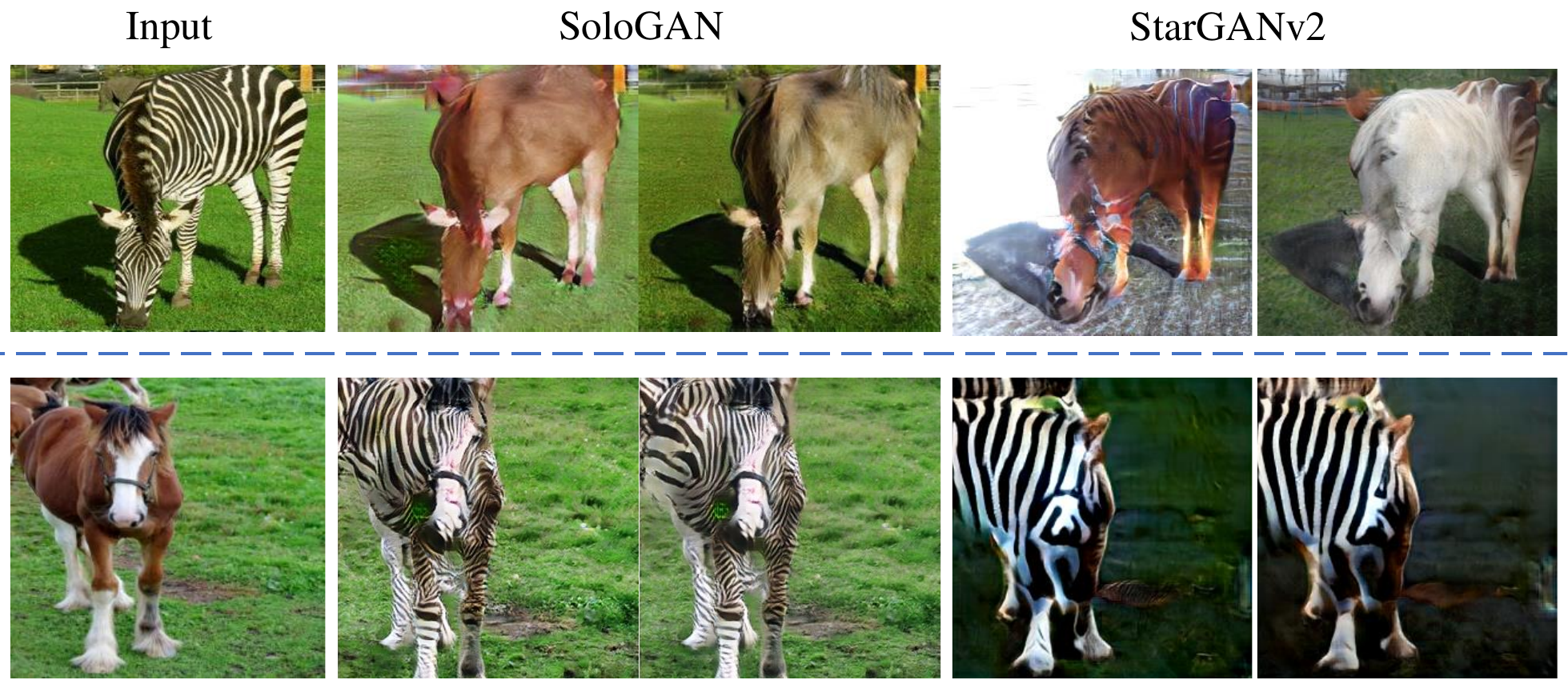}
    \caption{Qualitative comparison between SoloGAN and StarGANv2 on ${cat} \leftrightarrow {dog}$ (upper) and ${horse} \leftrightarrow {zebra}$ (bottom) datasets. }
\label{fig:add_horse2zebra}
\end{center}
\vspace{-0.4cm}
\end{figure}

We perform further empirical comparisons with StarGANv2~\cite{choi2020stargan} which is retrained with default settings from official implementation. Given the similar translation success rate between StarGANv2 and SoloGAN (99.96\% \emph{vs} 99.95\%), SoloGAN achieves slightly higher IS (1.025 \emph{vs} 1.051) and CIS (1.021 \emph{vs} 1.031) on ${cat}$$ \leftrightarrow$$ {dog}$ dataset, indicating the better potential of SoloGAN in generating more realistic samples than StarGANv2. 
As further evidenced by the visual comparisons in Fig.~\ref{fig:add_horse2zebra}, StarGANv2 totally ignores the grassland and completely fails to generate realistic and background-unchanged images on ${horse}$$ \leftrightarrow$$ {zebra}$.
Moreover, StarGANv2 is four times as larg as our SoloGAN (57.25M \emph{vs} 13.99M), which is impractical for mobile-setting devices.

\begin{figure}[t]
\begin{center}
    \includegraphics[width=0.32\textwidth]{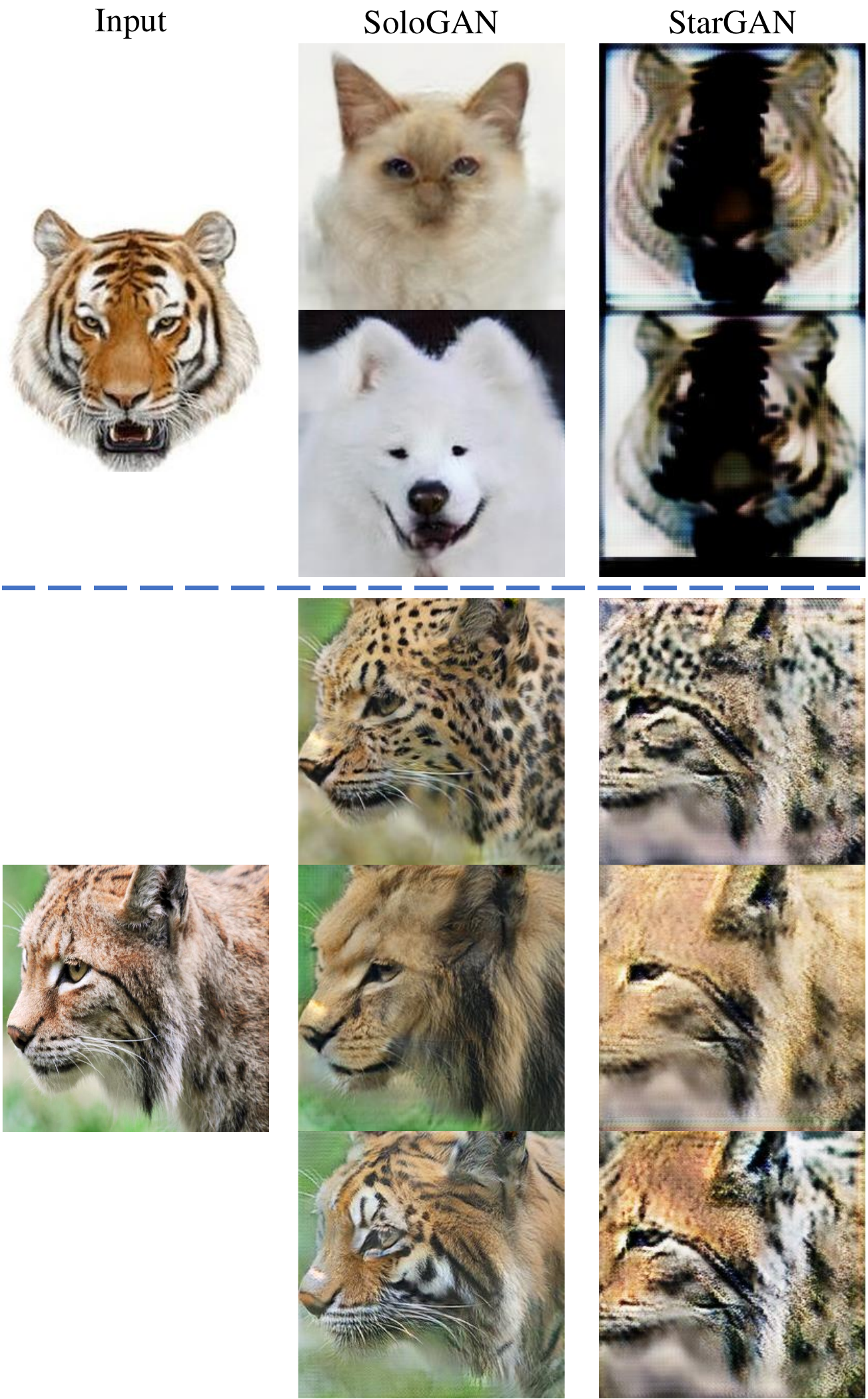}
    \caption{Qualitative comparison between SoloGAN and StarGAN on ${tiger} \rightarrow {cat, dog}$ (top) and ${bobcat} \rightarrow {leopard, lion, tiger}$ (bottom) datasets. In each dataset, first column shows input, and each subsequent column shows different domain outputs from a method.}
\label{fig:multi_domain}
\end{center}
\vspace{-0.5cm}
\end{figure}

In a word, our SoloGAN is a lightweight yet robust GAN across a wide range of datasets (especially those involving extreme shape variations and complex unchanged backgrounds), indicating its promising practical applicability.

\begin{table}[!htbp]
\centering
\caption{Translation success rates for multi-domain translation generated by SoloGAN and StarGAN. For fair comparison, we sample one image randomly from multimodal outputs obtained by SoloGAN given an input image. }
\label{tb:multi_domain_comparison}
\small
\begin{tabular}{c|c|c}
\toprule
Dataset & SoloGAN    & StarGAN~\cite{StarGAN} \\
\hline
$ cat \leftrightarrow dog \leftrightarrow tiger $ & \textbf{99.17}\% &39.00\%  \\
$leopard \leftrightarrow lion \leftrightarrow tiger \leftrightarrow bobcat$ & \textbf{92.08}\% & 41.25\% \\
\bottomrule
\end{tabular}
\vspace{-0.4cm}
\end{table}

\noindent{  \textbf{Multi-domain I2I translation.}
In addition to comparisons with multimodal methods between two domains, we provide a detailed discussion between our SoloGAN and StarGAN when the number of translation domains is higher than two. 
In this part, we randomly select one image from the multimodal outputs of our SoloGAN for each domain. 
Fig.~\ref{fig:multi_domain} shows the results generated by the SoloGAN and StarGAN models for multi-domain translation on the ${tiger}$ $\rightarrow {cat, dog}$ and ${bobcat}$ $\rightarrow {leopard, lion, tiger}$ datasets, where the corresponding translation success rate are given in Table~\ref{tb:multi_domain_comparison}.
Notably, it is difficult to evaluate the LPIPS, IS, CIS, and FID results, as StarGAN is a unimodal translation method. 
It can be observed that the StarGAN is ineffective in translating a tiger to cat or dog images (Fig.~\ref{fig:multi_domain}, right column), for which an effective model needs to account for significant shape changes. 
Nevertheless, the StarGAN has better performance on the ${bobcat}$$\rightarrow$${leopard, lion, tiger}$ dataset (Fig.~\ref{fig:multi_domain}, bottom) than on the ${tiger}$$\rightarrow$${cat, dog}$ dataset since the former requires less shape change and more focus on texture (such as strips and dots).  
The quantitative results are coincident with the qualitative results. SoloGAN achieved much higher translation success rates than StarGAN on both datasets, especially on the ${cat}$$\leftrightarrow {dog}$$\leftrightarrow{tiger}$ dataset (0.83\% $vs.$ 61\%). To summarize, the SoloGAN performs significantly better than the StarGAN model in dealing with multi-domain I2I translations. We argue that our disentangled representation of content and style and the novel discriminator are effective in generating realistic and target domain images.
}

\begin{figure*}[!htbp]
\centering
    \includegraphics[width=0.92\textwidth]{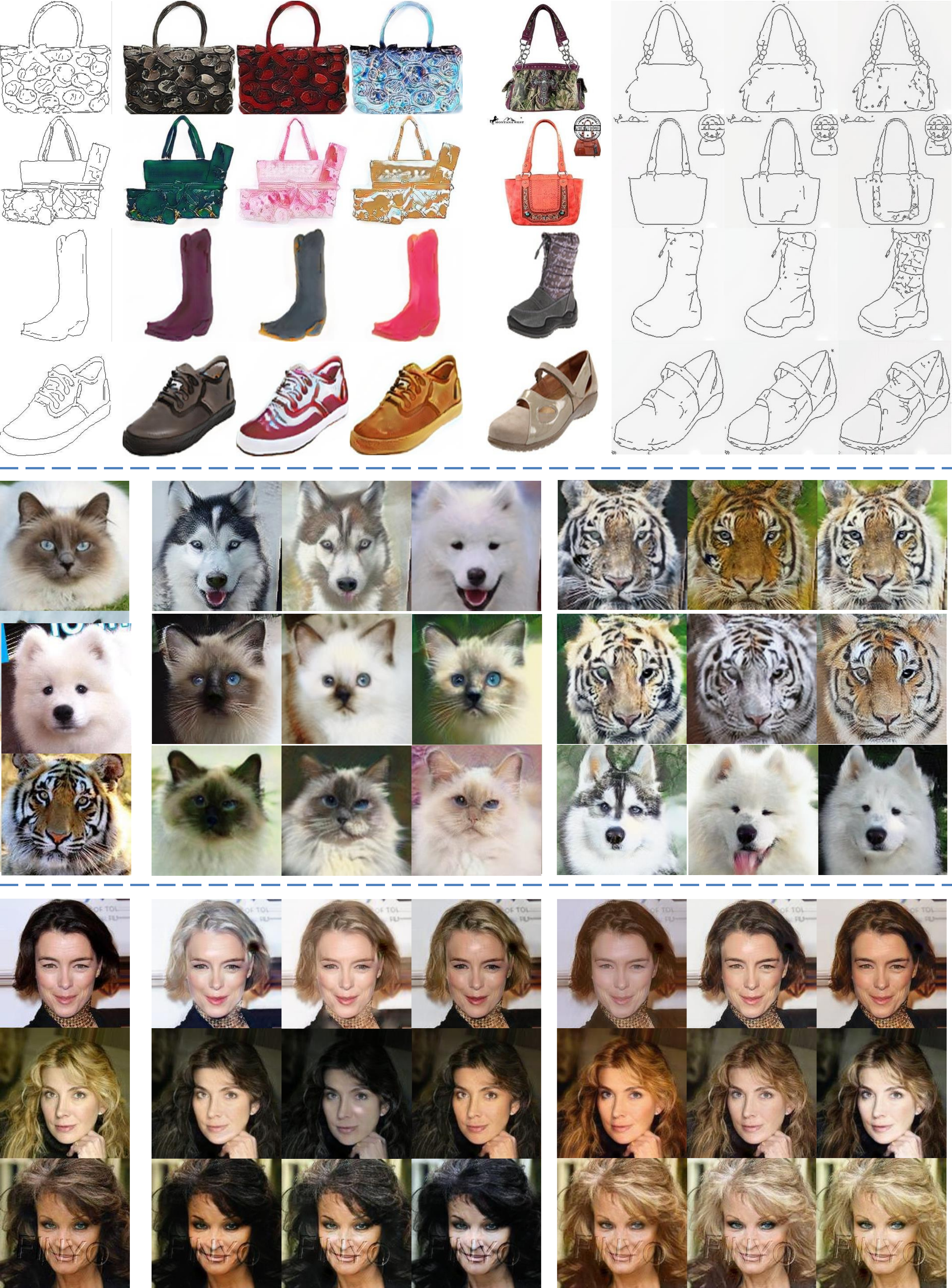}
   \caption{Translated results achieved by SoloGAN on a two-domain and two three-domain datasets. From top to bottom: multimodal sample outputs for the datasets of ${edges}\leftrightarrow {bags\&shoes}$, ${cat} \leftrightarrow {dog} \leftrightarrow {tiger}$, and ${black} \leftrightarrow {blond} \leftrightarrow {brown}$. }
\label{fig:two_domains_results}
\vspace{-0.3cm}
\end{figure*}

\begin{figure*}[!htbp]
\centering
    \includegraphics[width=0.81\textwidth]{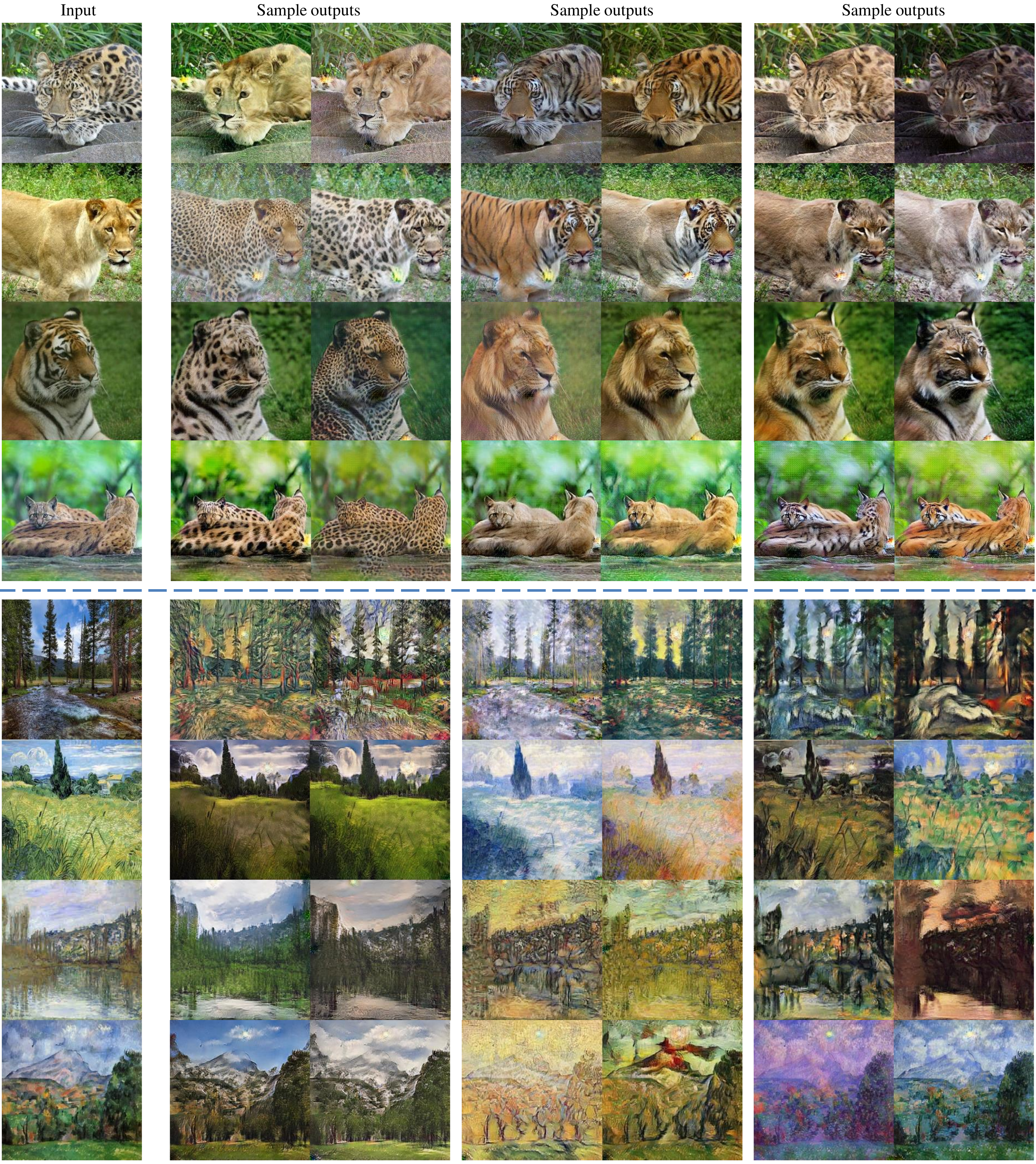}
      \caption{Translated results achieved by SoloGAN on two four-domain datasets. From top to bottom: multimodal sample outputs for the datasets of ${leopard} \leftrightarrow{lion} \leftrightarrow{tiger} \leftrightarrow{bobcat}$ and ${photo} \leftrightarrow{Van Gogh} \leftrightarrow{Monet} \leftrightarrow{Cezanne}$. Leftmost is input from one domain, while left column lists outputs translated from input into other three domains. For example, taking leopard input as first row, from left to right are generated lions, tigers, and bobcat, respectively.}
\label{fig:four_domains_results}
\vspace{-0.5cm}
\end{figure*}

\noindent{\textbf{More visual results}. We provide more visual results achieved by our SoloGAN to better understand its mechanism. It is necessary to claim again that there is only one single pair of generator and discriminator for each dataset, no matter how many image domains are involved. Additional translation results on two-domain datasets are presented in Fig.~\ref{fig:two_domains_results}. 
Apart from the results on the ${horse}$$ \leftrightarrow$$ {zebra}$ and ${day}$$\leftrightarrow$$ {night}$ datasets, we present the results on the ${summer}$$\leftrightarrow $${winter}$ and ${edges}$$ \leftrightarrow$$ {bags\&shoes}$ datasets.
Fig.~\ref{fig:two_domains_results} and Fig.~\ref{fig:four_domains_results} show some synthesized results over two three-domain and two four-domain datasets. It can be observed that all results are partly realistic and have good diversity. The stripes and dots of the translated leopards and tigers are different from each other in the dataset of ${leopard} \leftrightarrow lion \leftrightarrow tiger \leftrightarrow bobcat$.  
}

\begin{figure}[t]
\centering
\begin{center}
\begin{tabular}{ccc}
\includegraphics[width=1.7in]{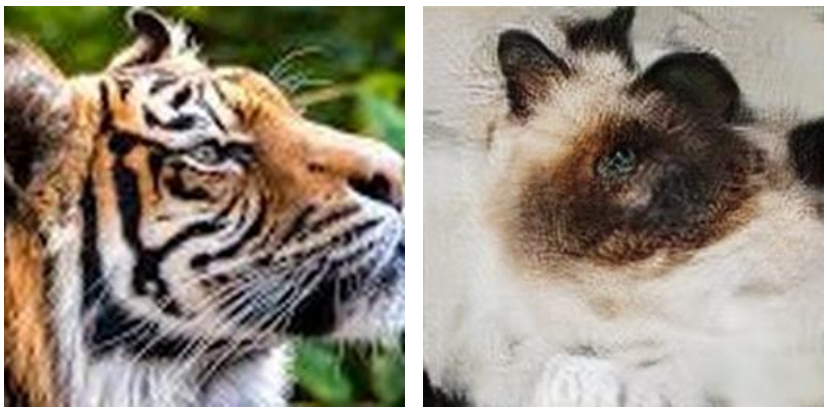} &
\includegraphics[width=1.7in]{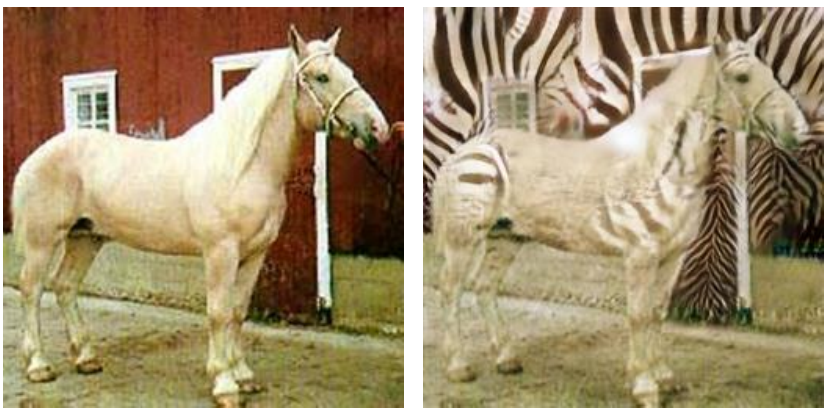} \\
                          (a) ${tiger} \rightarrow  {cat} $ & (b) $ horse \rightarrow zebra $ 
\end{tabular}
\end{center}
\vspace{-0.25cm}
  \caption{Typical failure cases of proposed SoloGAN. SoloGAN fails to translate a tiger with unusual face direction and an uncommon white horse in front of a building that has color similar to common horses into a cat and zebra, respectively.}
\label{fig:failure_cases}
\vspace{-0.52cm}
\end{figure}

\noindent{\textbf{Limitations}}. Just as did in the cases of other variants of GAN, the performance of SoloGAN can be limited by the images in the training set (e.g. number of images and poses). 
As shown by the example in Fig.~\ref{fig:failure_cases}(a), since there is no image in the ${cat}$ dataset having the same head pose as the tiger, SoloGAN fails to translate images in the target domain;
similarly, SoloGAN does not perform well when only a few training images of white horses are available (Fig.~\ref{fig:failure_cases}(b)).

\section{Conclusion}    \label{conclusion}

In this work, we propose an unpaired method for multi-domain multimodal  image-to-image translation, where a generative adversarial network--SoloGAN--is tailored.
In SoloGAN, the content/style encoder and the generator are shared among multiple domains, and a classification discriminator is tailored by combining with projection.
Experimental results have demonstrated the promising performance of SoloGAN in terms of both effectiveness and efficiency, especially for datasets involving complex image backgrounds or significant shape changes.


In the future, we will investigate the potentials of SoloGAN towards more challenging I2I translation datasets.
For example,  by pairing our proposed encoder and generator with different discriminators for multiple domains, SoloGAN can be potentially extended to deal with blended attribute I2I translation ~\cite{IntersectGAN}.


\bibliographystyle{IEEEtran}
\bibliography{IEEEabrv}


\end{document}